\documentclass[10pt,twocolumn,letterpaper]{article}

\usepackage{latex/cvpr}              

\usepackage{graphicx}
\usepackage{amsmath}
\DeclareMathOperator*{\argmin}{arg\,min}
\usepackage{amssymb}
\usepackage{booktabs}
\usepackage{bbding}
\usepackage{color}
\usepackage{algorithm}
\usepackage{algorithmicx}
\usepackage{algpseudocode}
\usepackage{makecell}
\usepackage{bbding}
\usepackage[misc]{ifsym}
\usepackage{times}
\usepackage{epsfig}
\usepackage[pagebackref,breaklinks,colorlinks]{hyperref}
\usepackage{multirow}
\usepackage{enumitem}
\usepackage[accsupp]{axessibility}

\definecolor{MyDarkBlue}{rgb}{0,0.08,1}
\definecolor{MyDarkGreen}{rgb}{0.02,0.6,0.02}
\definecolor{MyDarkRed}{rgb}{0.8,0.02,0.02}
\definecolor{MyDarkOrange}{rgb}{0.40,0.2,0.02}
\definecolor{MyPurple}{RGB}{111,0,255}
\definecolor{MyRed}{rgb}{1.0,0.0,0.0}
\definecolor{MyGold}{rgb}{0.75,0.6,0.12}
\definecolor{MyDarkgray}{rgb}{0.66, 0.66, 0.66}

\newcommand{\eqn}[1]{Eq.~(\ref{#1})}
\newcommand{\sect}[1]{Section~\ref{#1}}
\newcommand{\tab}[1]{Table~\ref{#1}}
\newcommand{\fig}[1]{Fig.~\ref{#1}}
\newcommand{\alg}[1]{Alg.~\ref{#1}}

\newcommand{\myparagraph}[1]{\vspace{-10pt}\paragraph{#1}}
\newcommand{\myitem}{\vspace{-5pt}\item}

\newcommand\blfootnote[1]{%
\begingroup
\renewcommand\thefootnote{}\footnote{#1}%
\addtocounter{footnote}{-1}%
\endgroup
}

\def\cvprPaperID{6337} 
\def\confName{CVPR}
\def\confYear{2022}

\begin{document}


\title{Continual Predictive Learning from Videos}

\author{
Geng Chen$^{1*}$\quad
Wendong Zhang$^{1*}$\quad
Han Lu$^{1}$\quad
Siyu Gao$^{1}$\quad
Yunbo Wang$^{1\dagger}$\\
Mingsheng Long$^{2}$\quad
Xiaokang Yang$^{1}$\\
$^1$MoE Key Lab of Artificial Intelligence, AI Institute, Shanghai Jiao Tong University\\
$^2$School of Software, BNRist, Tsinghua University\\
{\tt\small \{chengeng, diergent, yunbow\}@sjtu.edu.cn}
}

\maketitle

\begin{abstract}

Predictive learning ideally builds the world model of physical processes in one or more given environments. Typical setups assume that we can collect data from all environments at all times. In practice, however, different prediction tasks may arrive sequentially so that the environments may change persistently throughout the training procedure. Can we develop predictive learning algorithms that can deal with more realistic, non-stationary physical environments? In this paper, we study a new continual learning problem in the context of video prediction, and observe that most existing methods suffer from severe catastrophic forgetting in this setup. To tackle this problem, we propose the continual predictive learning (CPL) approach, which learns a mixture world model via predictive experience replay and performs test-time adaptation with non-parametric task inference. We construct two new benchmarks based on RoboNet and KTH, in which different tasks correspond to different physical robotic environments or human actions. Our approach is shown to effectively mitigate forgetting and remarkably outperform the na\"ive combinations of previous art in video prediction and continual learning. 

\end{abstract}
\section{Introduction}
\label{sec:intro}

\blfootnote{$^*$\ Equal contribution.}
\blfootnote{$^{\dagger}$\ Corresponding author: Yunbo Wang.}
\vspace{-10pt}

\begin{figure}[t]
    \centering
    \includegraphics[width=0.9\columnwidth]{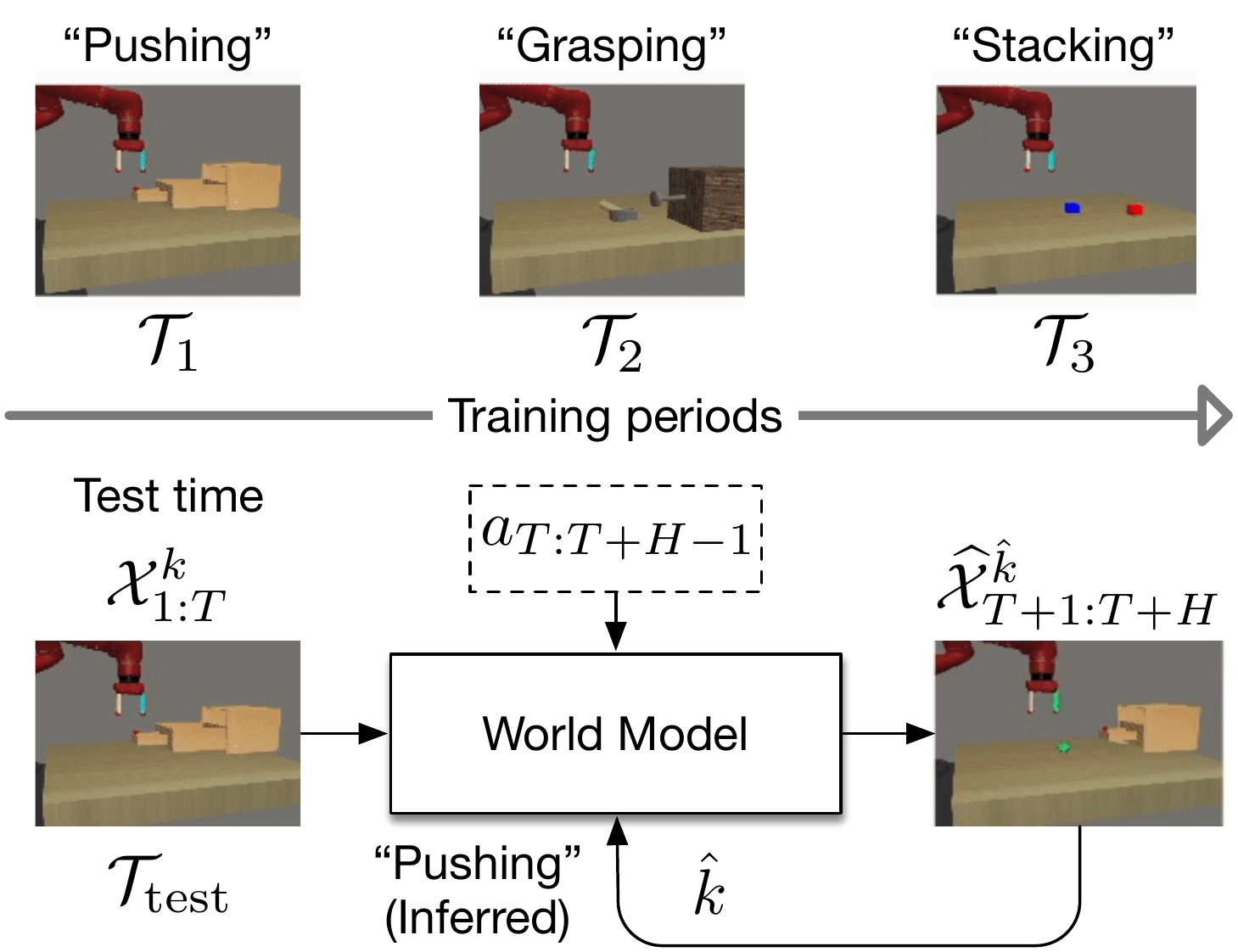}
    \vspace{-5pt}
    \caption{The new problem of continual predictive learning and the general framework of our approach at test time.}
    \label{fig:intro}
    \vspace{-10pt}
\end{figure}

Predictive learning is an unsupervised learning technique to build a world model of the environment by learning the consequences from historical observations, sequences of actions, and corresponding future observation frames.
The standard predictive learning setup is assumed to operate the model in a stationary environment with relatively fixed physical dynamics~\cite{srivastava2015unsupervised,shi2015convolutional,denton2018stochastic,guen2020disentangling}.
However, the assumption of stationarity does not always hold in more realistic scenarios, such as in the settings of continual learning (CL), where the model is learned through tasks that arrive sequentially. 
For example, in robotics (see \fig{fig:intro}), world models often serve as the representation learners of model-based control systems~\cite{ha2018recurrent,finn2017deep,hafner2019learning,hafner2020dream}, while the agent may be subjected to non-stationary environments in different training periods. 
Under these circumstances, it is not practical to maintain a single model for each environment or each task, nor is it practical to collect data from all environments at all times.
A primary finding of this paper is that most existing predictive networks~\cite{srivastava2015unsupervised,shi2015convolutional,denton2018stochastic,guen2020disentangling} cannot perform well when trained in non-stationary environments, suffering from a phenomenon known as catastrophic forgetting~\cite{goodfellow2013empirical}.

We formalize this problem setup as \textit{continual predictive learning}, in which the world model is trained in time-varying environments (\ie, ``tasks'' in the context of continual learning) with non-stationary physical dynamics. The model is expected to handle both newer tasks and older ones after the entire training phase (see \sect{sec:problem} for detailed setups). There are two major challenges.

\subsection{Covariate--Dynamics--Target Shift}
Unlike in the settings of \textit{domain-incremental} or \textit{class-incremental} CL for deterministic models, the world model, which can be viewed as a conditioned generative model, cannot assume a stationary distribution of training targets or fixed target space. 
Therefore, different from all previous CL problems, the unique challenge of continual predictive learning is due to \textit{the co-existence of three types of distribution shift, including the covariate shift in $P_X$, the target shift in $P_{Y}$, and the dynamics shift $P_{{Y}|{X}}$\footnote{In predictive learning settings, the input $X$ is in forms of sequential observation frames $\mathcal{X}_{1:T}$ and the training target $Y$ corresponds to future frames $\mathcal{X}_{T+1:T+H}$. We here skip the input action signals for simplicity.}.}
Notably, the covariate shift~\cite{shimodaira2000improving,sugiyama2007direct,gretton2009covariate,long2017deep,long2015learning,long2017conditional,liu2019transferable} and target shift~\cite{zhang2013domain,iyer2014maximum,lipton2018detecting,azizzadenesheli2019regularized,guo2020ltf} have been widely considered by existing methods, whereas the conditional distribution is typically assumed to be invariant. 
In our setup, however, the conditional distribution $P_{{Y}|{X}}$ corresponding to the spatiotemporal dynamics also changes over training periods. 
It significantly increases the probability of catastrophic forgetting in the world model.

To combat the dynamics shift, we first present a new world model that learns multi-modal visual dynamics of different tasks on top of task-specific latent variables.
Future frames are generated by drawing samples from learned mixture-of-Gaussian priors conditioned on a set of categorical task variables, and combining them with a deterministic component of future prediction (see \sect{sec:arch}).

Second, we specifically design a novel training scheme named \textit{predictive experience replay}. Like deep generative replay (DGR)~\cite{shin2017continual}, the proposed training method leverages a learned generative model to produce samples of previous tasks. Yet, in our approach, these samples are fed into the world model as the first frames to generate entire sequences, which can be reused as model inputs for rehearsal.
The world model alternates between (i) generating rehearsal data without backpropagating the gradients, (ii) regressing the facilitate future frames of previous tasks produced by the world model itself, and (iii) generating future frames from real data of the current task.
Another advantage of this training scheme is about the memory efficiency, as it only retains parts of low-dimensional action vectors in the buffer for action-conditioned predictive replay (see \sect{sec:replay}). 

\subsection{Task Inference: Coupled Forgetting Issues}

The second challenge in continual predictive learning is the task ambiguity at test time, which can greatly affect the prediction results. 
Unlike existing CL methods for fully generative models~\cite{shin2017continual,rao2019continual}, in our setup, the models are required not only to solve each task seen so far, but also to infer which task they are presented with. 
A na\"ive solution is to infer the task using another neural network.
However, due to the inevitable forgetting issue of the task inference model itself, coupled with that of the world model, this method is unlikely to perform well.
In \sect{sec:inference}, we propose the non-parametric task inference strategy, which overcomes the intrinsic nature of forgetting of a deterministic model. 
We also present a self-supervised, test-time training process that recalls the pre-learned knowledge of the inferred task through one or several online adaptation steps.

We construct two new benchmarks for continual predictive learning based on real-world datasets, RoboNet~\cite{DBLP:conf/corl/DasariETNBSSLF19} and KTH~\cite{DBLP:conf/icpr/SchuldtLC04}, in which different tasks correspond to different physical robotic environments or human actions. Our CPL approach is shown to effectively avoid forgetting and remarkably outperform the straightforward combinations of previous art in video prediction and continual learning.

\section{Problem Setup}
\label{sec:problem}

Unlike existing predictive learning approaches, we consider to learn a world model ($\mathcal{M}$) in non-stationary environments (\ie, the evolution of tasks), such that 
\begin{equation}
\label{eq:setup}
    \widehat{\mathcal X}_{T+1:T+H} \sim \mathcal{M}(\mathcal X_{1:T}, a_{T:T+H-1}, \hat{k}),
\end{equation}
where $\mathcal X_{1:T}$ and $\mathcal X_{T+1:T+H}$ are respectively the observed frames and future frames to be predicted. 
The task index $k$ is known at training, but not observed at test. It requires our approach not only to solve each task seen so far, but also to infer which task it is presented with, denoted as $\mathcal{T}_{\hat{k}}$.
Here, $a_{T:T+H-1}$ is the optional inputs of action signals when $\mathcal{M}$ is learned for vision-based robot control, as in the action-conditioned video prediction experiments in this paper.
Formally, continual predictive learning assumes that: 
\begin{equation}
\small
\begin{split}
\label{eq:shift}
    \text{Covariate shift:} & \ P(\mathcal X_{1:T}^k) \neq P(\mathcal X_{1:T}^{k+1}) \\
    \text{Dynamics shift:} & \ P(\mathcal X_{T+1:T+H}^k | \mathcal X_{1:T}^k) \neq P(\mathcal X_{T+1:T+H}^{k+1} | \mathcal X_{1:T}^{k+1}) \\
    \text{Target shift:} & \ P(\mathcal X_{T+1:T+H}^k) \neq P(\mathcal X_{T+1:T+H}^{k+1}),
\end{split}
\end{equation}
where we leave out $a_{T:T+H-1}$ for simplicity in the conditional distribution of visual dynamics.
The setup is in part similar to class-incremental CL for supervised tasks that assumes $P(\mathcal X^k) \neq P(\mathcal X^{k+1})$, $\{\mathcal Y^k\} = \{\mathcal Y^{k+1}\}$, $P(\mathcal Y^k) \neq P(\mathcal Y^{k+1})$.
$\{\mathcal Y^k\}$ denotes a constant label set for discriminative models. 
In contrast, continual predictive learning does not assume a fixed target space, and therefore may have more severe catastrophic forgetting issues.

\section{Approach}
\newcommand{\x}{{\mathcal{X}}}
\newcommand{\z}{\textbf{z}}
\newcommand{\M}{{\mathcal{M}}}
\newcommand{\G}{{\mathcal{G}}}

\begin{figure*}[t]
\centering
\includegraphics[width=1.0\linewidth]{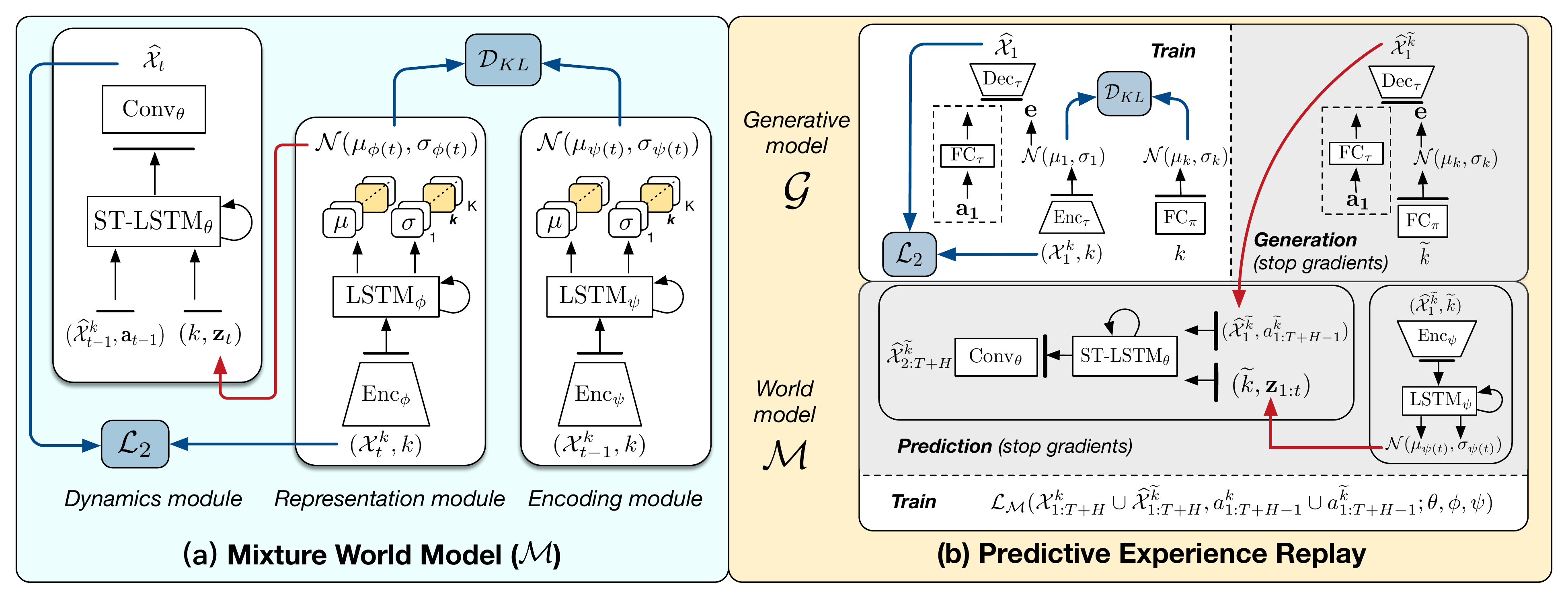}
\vspace{-20pt}
\caption{The overall network architecture of the \textit{mixture world model} and the \textit{predictive experience replay} training scheme in the proposed CPL method. (a) The world model learns representations in the forms of mixture-of-Gaussians based on categorical task variables. (b) As for the predictive experience replay, the world model ($\M$) interacts with the initial-frame generative model ($\G$).
In this replay stage, we first use $\G$ to generate the first frames of previous tasks without backpropagating the gradients, then use $\M$ to predict the corresponding future frames, and finally combine the rehearsal data and real data to jointly train $\M$ and $\G$.}
\label{fig:method}
\vspace{-5pt}
\end{figure*}

In this section, we present the new continual predictive learning (CPL) approach, which first mitigates catastrophic forgetting within the world model from two aspects: 
\begin{itemize}[leftmargin=*]
    \myitem \textbf{Mixture world model:} A recurrent network that captures multi-modal visual dynamics. Unlike existing models \cite{denton2018stochastic,hafner2020dream}, the learned task-specific priors are in forms of mixture-of-Gaussians to overcome dynamics shift. 
    \myitem \textbf{Predictive experience replay:} A new rehearsal-based training scheme that combats the forgetting within the world model and is efficient in memory usage.
\end{itemize}
\vspace{-5pt}

To cope with the challenge of task ambiguity when testing the world model in an unknown task, we propose:
\begin{itemize}[leftmargin=*]
    \myitem \textbf{Non-parametric task inference:} Instead of using any parametric task inference model that may introduce extra forgetting issues, we use a trial-and-error strategy over the task label set to determine the present task. 
\end{itemize}

\subsection{Mixture World Model}
\label{sec:arch}

The world model $\M$ considers a new remedy to catastrophic forgetting from the perspective of spatiotemporal representation.
As mentioned above, the forgetting problem within existing world models \cite{denton2018stochastic,hafner2020dream} is mainly caused by the covariate-dynamics-target shift in time-varying environments. 
Therefore, the key idea of the proposed world model in CPL is to use mixture-of-Gaussian variables to capture the multi-modal distribution of visual dynamics in the latent space, as well as that of spatial appearance in the input/output observation space.
Accordingly, as shown in \fig{fig:method}\textcolor{red}{(a)}, the world model consists of three components:
\begin{equation}
\begin{split}
    \text{Representation module:} & \quad \z_{t} \sim q_{\phi}({\x}_{1:t}^k, k) \\
    \text{Encoding module:} & \quad \hat{\z}_{t} \sim p_{\psi}({\x}_{1:t-1}^k, k) \\
    \text{Dynamics module:} & \quad \widehat{\x}_{t} =  p_{\theta}(\x_{1:t-1}^k,a_{1:t-1},\z_{1:t}, k).\\
\end{split}
\end{equation}
The representation module infers the latent state $\z_{t}$ from the target frames. It takes as input the categorical task variable $k \in 
\{1,\ldots,K\}$ to cope with the target shift in continual predictive learning scenarios. 
The encoding module corresponds to the covariate shift and dynamically maps the input frames to $\hat{\z}_{t}$ in the same latent subspace as $\z_{t}$.
The dynamics module learns the deterministic transition component from inputs to prediction targets.
It responds to multi-modal spatiotemporal dynamics by taking as input the task-specific latent variable $\z_{t}$. 
All components are implemented as neural networks, in which the dynamics module is particularly composed of stacked ST-LSTM layers~\cite{wang2017predrnn}.

The task-specific latent representation $\z_{t}$ is drawn from a mixture-of-Gaussian distribution, inspired by existing unsupervised learning methods that use Gaussian Mixture priors for variational autoencoders \cite{dilokthanakul2016deep,jiang2017variational,rao2019continual}. 
Our mixture world model is an early work that uses this representation form to model the multi-modal priors in spatiotemporal dynamics. 
Specifically, for each task, the representation module and the encoding module are both conditioned on the present task label. 
They are jointly trained to learn the posterior and prior distribution of $\z_{t}$ by optimizing the Kullback-Leibler divergence.
%
At task $\mathcal T_k$, the objective function ${\mathcal{L}}_{\M}^k(\x_{1:T+H}^k,a_{1:T+H-1}^k)$ combines the KL loss with the reconstruction loss:
\begin{equation}
\begin{split}
    \sum\limits_{t=2}^{T+H} & \big[ \mathbb{E}_{q(\z_{1:t} \, | \, {{\mathcal X}_{1:t}^k,k})}\log p({\mathcal X}_{t}^k \, | \, {\mathcal X}_{1:t-1}^k,a_{1:t-1}^k,\z_{1:t},k)\\ 
    &-\alpha D_{KL}(q(\z_{t} \, | \, {\mathcal X}_{1:t}^k,k) \, || \, p(\hat{\z}_{t} \, | \,  {\mathcal X}_{1:t-1}^k,k))\big], \\
\end{split}
\label{eq:KL}
\end{equation}
where $\alpha$ is set to $10^{-4}$ in our experiments. In the test phase, we discard the representation module $q_{\phi}$ and only use the encoding module $p_{\psi}$ to sample task-specific latent variables for frames generation.

\subsection{Predictive Experience Replay}
\label{sec:replay}

The two main challenges in typical CL setups are catastrophic forgetting and memory limitation.
Due to the co-existence of covariate shift, target shift, and dynamics shift, these challenges become even more urgent in the context of continual predictive learning based on video data.
One common way to tackle these challenges is generative replay \cite{shin2017continual,rao2019continual}, which considers using a generative model to produce samples of previous tasks. 
However, the generative replay method cannot be used directly in our setup, as it is extremely difficult to generate a valid video sequence using a generative model alone.

Therefore, we propose the predictive experience replay, which firmly combines an initial-frame generative model ($\G$), which learns to generate the first frame of videos at previous tasks given the task labels, with the world model ($\M$).
To counter the coviariate shift of image appearance in non-stationary environments, $\G$ also uses learnable mixture-of-Gaussian latent priors, denoted by $\boldsymbol{e}$.
As shown in \fig{fig:method}\textcolor{red}{(b)}, for each previous task $\mathcal T_{\tilde{k}}$, we first use $\G$ to generate the first frames of the rehearsal video sequences, and then use $\M$ to predict the corresponding future frames. 
Finally, we mix the rehearsal sequences at previous tasks and real sequences at the present task $\mathcal T_k$ to train $\G$ and $\M$ in turn. We summarize the training procedure in \alg{alg:train}.
The predictive experience replay is different from all existing generative replay methods because the world model plays a key role in the rehearsal process.

\begin{algorithm}[t] 
  \caption{Predictive experience replay}  
  \label{alg:train}  
  \begin{algorithmic}[1]
    \Require  
    Training data $\{ \x_{1:T+H}^{k}\}_{k=1}^K $, $\{a_{1:T+H-1}^{k}\}_{k=1}^K $
    \Ensure  World model $\M$, generative model $\G$ 
    \State Train $\M$ at $\mathcal{T}_1$ according to \eqn{eq:KL} 
    \State Train $\G$ at $\mathcal{T}_1$ according to \eqn{eq:G_loss} with $k=1$
    \For{$k=2, \ldots, K$}
        \State \textit{\# Replay video sequences (skip the batch size)}
        \For{$\tilde{k}=1, \ldots, k-1$}
            \State $\widehat{\x}^{\tilde{k}}_{1} \leftarrow \G(a^{\tilde{k}}_{1},{\tilde{k}})$
            \State $\widehat{\x}_{2:T+H}^{\tilde{k}} \leftarrow \M(\widehat{\x}^{\tilde{k}}_{1},a_{2:T+H-1}^{\tilde{k}},{\tilde{k}})$
        \EndFor
         \State \textit{\# Mix  replayed data at $\mathcal{T}_{1:k-1}$ and real data at $\mathcal{T}_k$}
        \State $ (\widehat{\x}_{1:T+H}^{1:k-1},a_{1:T+H-1}^{1:k-1})  \cup (\x_{1:T+H}^k,a_{1:T+H-1}^k)$
        \State Train $\M$  according to \eqn{eq:M_loss}
        \State Train $\G$ according to \eqn{eq:G_loss}
    \EndFor
  \end{algorithmic}  
\end{algorithm}

%
%

In particular, for action-conditioned predictive learning scenarios, we maintain a buffer to keep parts (${\tiny{\sim}}7\%$) of the low-dimensional action sequences from previous tasks.
During predictive experience replay, we first sample an action sequence from the buffer $a_{1:T+H-1}^{\tilde{k}}$ at a previous task $\mathcal T_{\tilde{k}}$.
We feed the initial action $a_1^{\tilde{k}}$ and the task label $\tilde{k}$ into $\G$ to ensure that the generated first frame $\widehat{\x}_1^{\tilde{k}}$ is valid for robot control, and perform $\M$ to produce predictive replay results $\widehat{\x}_{2:T+H}^{\tilde{k}}$ given $\widehat{\x}_1^{\tilde{k}}$ and $a_{2:T+H-1}^{\tilde{k}}$.
In predictive experience replay, we train the world model $\M$ at $\mathcal{T}_k$ by minimizing
\begin{equation}
\begin{split}
    {\mathcal{L}}_{\M} = &  \sum\limits_{\tilde{k}=1}^{k-1} {\mathcal{L}}_{\M}^{\tilde{k}}(\widehat{\x}_{1:T+H}^{\tilde{k}},a_{1:T+H-1}^{\tilde{k}}) \\
    & + {\mathcal{L}}_{\M}^k(\x_{1:T+H}^k,a_{1:T+H-1}^k). \\
\end{split}
\label{eq:M_loss}
\end{equation}
The objective function of the initial-frame generative model $\G$ can be written as
\begin{equation}
\begin{split}
    {\mathcal L}_{\G} & = \ \mathbb{E}_{q(\boldsymbol{e} \, | \, {{\x}_{1}^k,k})} \log p({\x}_{1}^k \, | \, \boldsymbol{e},a_1^k,k) \\
    & \quad - \beta D_{KL}(q(\boldsymbol{e} \, | \, {\x}_{1}^k, k) \, || \, p(\hat{\boldsymbol{e}} \, | \, k)) \\
    & + \sum\limits_{\tilde{k}=1}^{k-1} \big[ \mathbb{E}_{q(\boldsymbol{e} \, | \, {\widehat{{\x}}_{1}^{\tilde{k}},{\tilde{k}}})} \log p(\widehat{{\x}}_{1}^{\tilde{k}} \, | \, \boldsymbol{e},a_1^{\tilde{k}},{\tilde{k}}) \\
    & \quad - \beta D_{KL}(q(\boldsymbol{e} \, | \, \widehat{{\x}}_{1}^{\tilde{k}}, {\tilde{k}}) \, || \, p(\hat{\boldsymbol{e}} \, | \, {\tilde{k}})) \big],\\
\end{split}
\label{eq:G_loss}
\end{equation}
where the reconstruction loss is in an $\ell_2$ form and $\beta$ is set to $10^{-4}$ through empirical grid search.

\subsection{Non-Parametric Task Inference}
\label{sec:inference}

\begin{algorithm}[t] 
  \caption{Testing procedure}  
  \label{alg:test}  
  \begin{algorithmic}[1]
    \Require  
     Observation frames $\x_{1:T}$, optional actions $a_{1:T+H}$
    \Ensure  Predicted future frames $\widehat{\x}_{T+1:T+H}$
    \State \textit{\# Non-parametric task inference}
    \For{$k=1, \ldots, K$}
        \State $\widehat{\x}_{T/2+1:T}^k \leftarrow \M( \x_{1:T/2}, a_{1:T-1}, k)$
    \EndFor
    \State $\hat{k} = \argmin_{k \in \{1,\ldots,K \}} \sum_{t=T/2+1}^{T}(\x_{t}-\widehat{\x}_{t}^k)^2$
    \State \textit{\# Test-time adaptation (optional)}
    \State Optimize $\mathcal M$ with ${\mathcal L}_{\M}^{\hat{k}}({\mathcal X}_{1:T},a_{1:T-1})$
    \State \textit{\# Model deployment}
    \State $\widehat{\mathcal X}_{T+1:T+H} \leftarrow \M(\x_{1:T}, a_{1:T+H-1}, \hat{k})$
  \end{algorithmic}  
\end{algorithm}

In the mixture world model, the task label has a significant impact on the learned priors and corresponding prediction results. 
Since it is unknown at test time, it can only be inferred from the input observation sequences, \ie, video classification.
However, existing video classification models tend to underperform in the domain-incremental CL setting, which will magnify the catastrophic forgetting problem jointly trained with the world model.
To avoid the inherent forgetting issue of model-based task inference, we propose a new non-parametric method that only exploits the learned mixture world model to make task inference.

\begin{table*}[t]
\centering
\small
    \begin{tabular}{|l|cc|cc|}
    \hline
    \multirow{2}{0cm}{Method}  & \multicolumn{2}{c|}{Action-conditioned} & \multicolumn{2}{c|}{Action-free} \\
    & PSNR$^{\uparrow}$ & SSIM$^{\uparrow}$ ($\times 10^{-2}$)  & PSNR$^{\uparrow}$ & SSIM$^{\uparrow}$ ($\times 10^{-2}$)  \\
    \hline
    \hline
    SVG~\cite{denton2018stochastic} &18.72 $\pm$ 0.61  & 68.59 $\pm$ 2.22 & 18.92 $\pm $0.51   &  68.08 $\pm$ 2.20      \\
    PredRNN~\cite{wang2017predrnn}  & 19.45 & 66.38  & 19.56    & 69.92        \\
    PhyDNet~\cite{guen2020disentangling} & 19.60 & 68.68 & 21.00 & 75.47 \\
    \hline
    PredRNN + LwF~\cite{li2017learning}& 19.10 & 64.73 & 19.79 & 71.43  \\
    PredRNN + EWC~\cite{kirkpatrick2017overcoming} &21.15&74.72&21.15&78.02\\
    CPL-base + EWC~\cite{kirkpatrick2017overcoming}   & 21.29 $\pm$ 0.30 & 75.16 $\pm$ 0.98 & 21.38 $\pm$ 0.18     & 76.68 $\pm$ 0.69    \\
    \hline
    CPL-base & 19.36 $\pm$ 0.00  & 63.57 $\pm$ 0.00 & 20.15 $\pm$ 0.02      & 71.15 $\pm$ 0.08            \\
    CPL-full & \textbf{23.26 $\pm$ 0.10}  & \textbf{80.72 $\pm$ 0.23} & \textbf{22.48 $\pm$ 0.03}     & \textbf{78.84 $\pm$ 0.07}  \\
    \hline
    CPL-base (Joint training)  & 24.64 $\pm$ 0.01  & 83.73 $\pm$ 0.00 &  22.56 $\pm$ 0.01    & 79.57 $\pm$ 0.02  \\
    \hline
    \end{tabular}
    \caption{Quantitative results of continual predictive learning on the RoboNet benchmark in both action-conditioned and action-free setups. (\textbf{Lines 1-3}) Existing video prediction models with i.i.d. assumption. 
    (\textbf{Lines 4-6}) Combinations of predictive models and continual learning approaches. 
    (\textbf{Lines 7-8}) Our predictive model based on learned mixture-of-Gaussian priors, and the the entire CPL with predictive experience replay and non-parametric task inference.  
    (\textbf{Line 9}) A baseline model jointly trained on all tasks throughout the training procedure, whose results can be roughly viewed as the upper bound of our approach.
	}
	\label{tab:robonet}
 	\vspace{-5pt}
\end{table*}

More precisely, as shown in \alg{alg:test}, we feed the first half of each input sequence into the world model $\x_{1:T/2}$, along with a hypothetic task label $k$.
We then enumerate each task label $k \in \{1,\ldots,K\}$ and evaluate the outputs of the world model on the remaining frames of the input sequence $\x_{T/2+1:T}$.
Finally, we choose the task label $\hat{k}$ that leads to the best prediction quality.

In addition to using $P(\x_{T/2+1:T} | \x_{1:T/2})$ to perform task inference, we also use this self-supervision for test-time adaptation, which allows the model to continue training after deployment. 
Test-time adaptation effectively recalls the pre-learned knowledge in the inferred task ${\mathcal{T}}_{\hat{k}}$ through one-step (or few-steps) online optimization, thus further alleviating the forgetting problem.

\section{Experiment}


\subsection{Experimental Setup}

\paragraph{Benchmarks.}
We quantitatively and qualitatively evaluate CPL on  the following two real-world datasets:
\begin{itemize}[leftmargin=*]
    \myitem \textbf{RoboNet} \cite{DBLP:conf/corl/DasariETNBSSLF19}. The RoboNet dataset contains action-conditioned videos of robotic arms interacting with a variety of objects in various environments. We divide the whole dataset into four continual learning tasks according to the environments (\ie, \textit{Berkeley} $\rightarrow$ \textit{Google} $\rightarrow$ \textit{Penn} $\rightarrow$ \textit{Stanford}). For each task, we collect about $3{,}840$ training sequences and $960$ testing sequences.
    \myitem \textbf{KTH action}\cite{DBLP:conf/icpr/SchuldtLC04}. This dataset contains gray-scale videos which include $6$ types of human actions. We directly use the action labels to divide the dataset into $6$ tasks (\ie, \textit{Boxing} $\rightarrow$ \textit{Clapping} $\rightarrow$ \textit{Waving} $\rightarrow$ \textit{Walking} $\rightarrow$ \textit{Jogging} $\rightarrow$ \textit{Running}). For each task, we collect about $1{,}500$ training sequences and $800$ testing sequences in average. 
\end{itemize}
\vspace{-5pt}
We define the task orders by random sampling, and without loss of generality, our approach is effective to any task orders (see \sect{sec:ablation}). More experimental configurations and the implementation details can be found in the Supplementary Material.

\myparagraph{Evaluation criteria.}
We adopt SSIM and PSNR from previous literature~\cite{denton2018stochastic,wang2017predrnn} to evaluate the prediction results. We run the continual learning procedure $10$ times and report the mean results and standard deviations in the two metrics. 

\myparagraph{Compared methods.}
We compare CPL with the following baselines and existing approaches:
\begin{itemize}[leftmargin=*]
    \myitem \textbf{CPL-base}: A baseline model that excludes the new components of Gaussian mixtures, predictive replay, and task inference. 
    \myitem \textbf{PredRNN} \cite{wang2017predrnn}, \textbf{SVG} \cite{denton2018stochastic}, \textbf{PhyDNet} \cite{guen2020disentangling}: Video prediction models focused on stochastic, deterministic, and disentangled dynamics modeling respectively.
    \myitem \textbf{LwF} \cite{li2017learning}: It is a distillation-based CL method built on the memory state of PredRNN \cite{wang2017predrnn}.
    \myitem \textbf{EWC} \cite{kirkpatrick2017overcoming}: It constrains the parameters of PredRNN and CPL-base on new tasks with additional loss terms. 
\end{itemize}
\vspace{-5pt}

 

\subsection{RoboNet Benchmark}
We first evaluate CPL on the real-world RoboNet benchmark, in which different continual learning tasks are divided by laboratory environments.
We conduct both action-conditioned and action-free video prediction on RoboNet.
The former follows the common practice~\cite{babaeizadeh2018stochastic,DBLP:conf/cvpr/WuNM0F21} to train the world model to predict $10$ frames into the future from $2$ observations and corresponding action sequence at the $11$ time steps. 
For the action-free setup, we use the first $5$ frames as input to predict the next $10$ frames. 

\begin{figure}[t]
\centering
\includegraphics[width=1.0\linewidth]{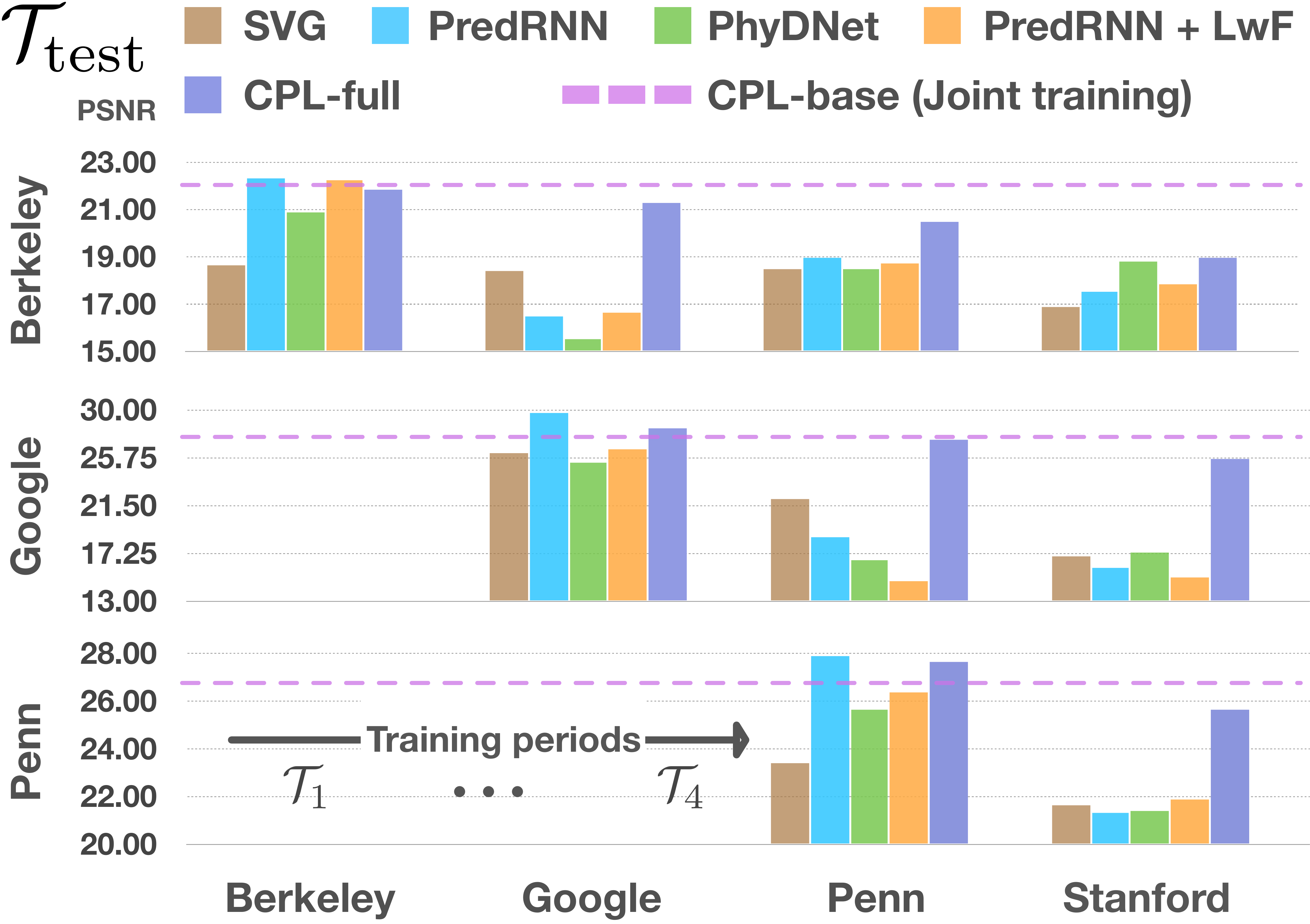}
\vspace{-15pt}
\caption{Results on the action-conditioned RoboNet benchmark. The horizontal axis represents the sequential training process, and the vertical axes represent test results on particular tasks after each training period. The purple dashed line indicates the results of the baseline model jointly trained on all tasks.
}
\vspace{-5pt}
\label{fig:robonet_action_cond}
\end{figure}

\myparagraph{Quantitative comparison.} 
\tab{tab:robonet} gives the quantitative results on RoboNet, in which the models are evaluated on the test sets of all $4$ tasks after the training period on the last task.
We have the following findings here.
\textbf{First}, CPL outperforms existing video prediction models by a large margin.
For instance, in the action-conditioned setup, it improves SVG in PSNR by $24.3\%$, PredRNN by $19.6\%$, and PhyDNet by $18.7\%$. 
\textbf{Second}, CPL generally performs better than previous continual learning methods (\ie, LwF and EWC) combined with video prediction backbones. 
Note that a na\"ive implementation of LwF on top of PredRNN even leads to a negative effect on the final results.
\textbf{Third}, by comparing CPL-full (our final approach) with CPL-base (w/o Gaussian mixture latents, predictive experience replay, or non-parametric task inference), we can see that the new technical contributions have a great impact on the performance gain.
We provide more detailed ablation studies in \sect{sec:ablation}.
\textbf{Finally}, CPL is shown to effectively ease catastrophic forgetting by approaching the results of jointly training the world model on all tasks in the i.i.d. setting ($23.26$ vs. $24.64$ in PSNR). 
Apart from the average scores for all tasks, in \fig{fig:robonet_action_cond}, we provide the test results on particular tasks after individual training periods.
As shown in the bar charts right to the main diagonal, CPL performs particularly well on previous tasks, effectively alleviating the forgetting issue. 
Please refer to the Supplementary Material for detailed comparison results.

\begin{figure}[t]
\centering
\includegraphics[width=1.0\linewidth]{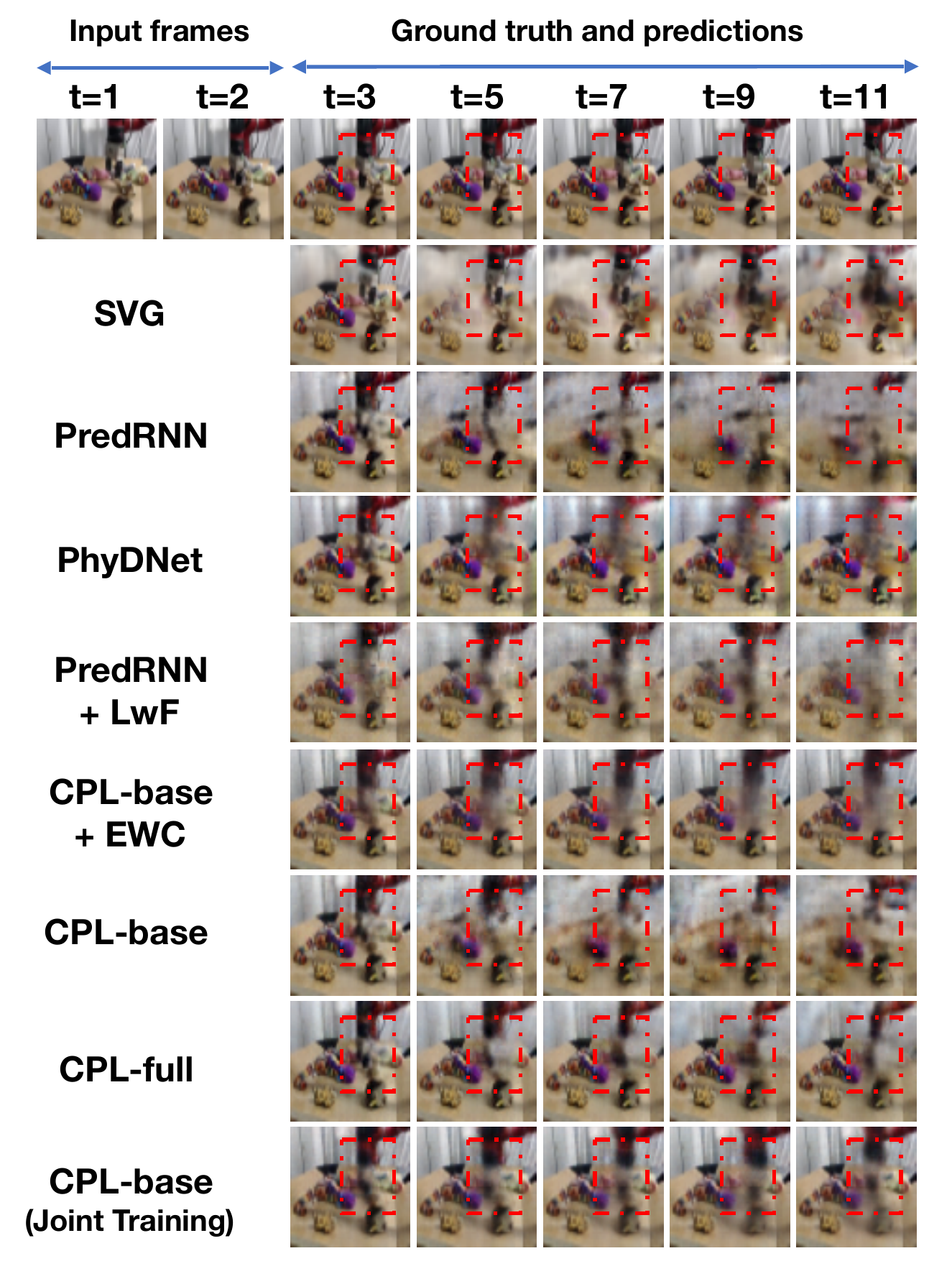}
\vspace{-25pt}
\caption{Showcases of action-conditioned video prediction in the first environment of RoboNet (\ie, \textit{Berkeley}) after training the models in the last environment (\ie, \textit{Stanford}).}
\label{fig:robonet_qualitative}
\end{figure}

\myparagraph{Qualitative comparison.}
\fig{fig:robonet_qualitative} provides the qualitative comparisons on the action-conditioned RoboNet benchmark.
Specifically, we use the final models after the training period of the last task to make predictions on the first task.
We can see from these demonstrations that our approach is more accurate in predicting both future dynamics of the objects as well as the static information of the scene.
In contrast, the predicted frames by PredRNN+LwF and CPL-base+EWC suffer from severe blur effect in the moving object or the static (but complex) background, indicating that directly combining existing CL algorithms with the world models cannot effectively cope with the dynamics shift in highly non-stationary environments.

\subsection{KTH Benchmark}

\begin{table}[t]
\small
    \centering
    \begin{tabular}{|l|cc|}
    \hline
    Method       & PSNR & SSIM ($\times 10^{-2}$) \\
    \hline
    \hline
    SVG~\cite{denton2018stochastic}          &   22.20 $\pm$ 0.02   &         69.23 $\pm$ 0.01       \\
    PredRNN~\cite{wang2017predrnn}                    &  23.27   &  70.47                    \\
    PhyDNet~\cite{guen2020disentangling}                    &  23.68   &  72.97                    \\
    \hline
    PredRNN + LwF~\cite{li2017learning} &  24.25    &     70.93                 \\
    CPL-base + EWC~\cite{kirkpatrick2017overcoming} &    24.32 $\pm$ 0.15  &         69.02 $\pm$ 0.48            \\
    \hline
    CPL-base     & 22.96 $\pm$ 0.05    &  68.98 $\pm$ 0.02                         \\
    CPL-full         &   \textbf{29.12 $\pm$ 0.03}  &  \textbf{84.50 $\pm$ 0.04}           \\
    \hline
    CPL-base (Joint train)     & 28.12 $\pm$ 0.01   &    82.16 $\pm$ 0.00   \\
    \hline
    \end{tabular}
    \vspace{-5pt}
    \caption{Quantitative results on the KTH benchmark.
	}
	\label{tab:kth}
	\vspace{-5pt}
\end{table}

\begin{figure}[t]
\centering
\includegraphics[width=1.0\linewidth]{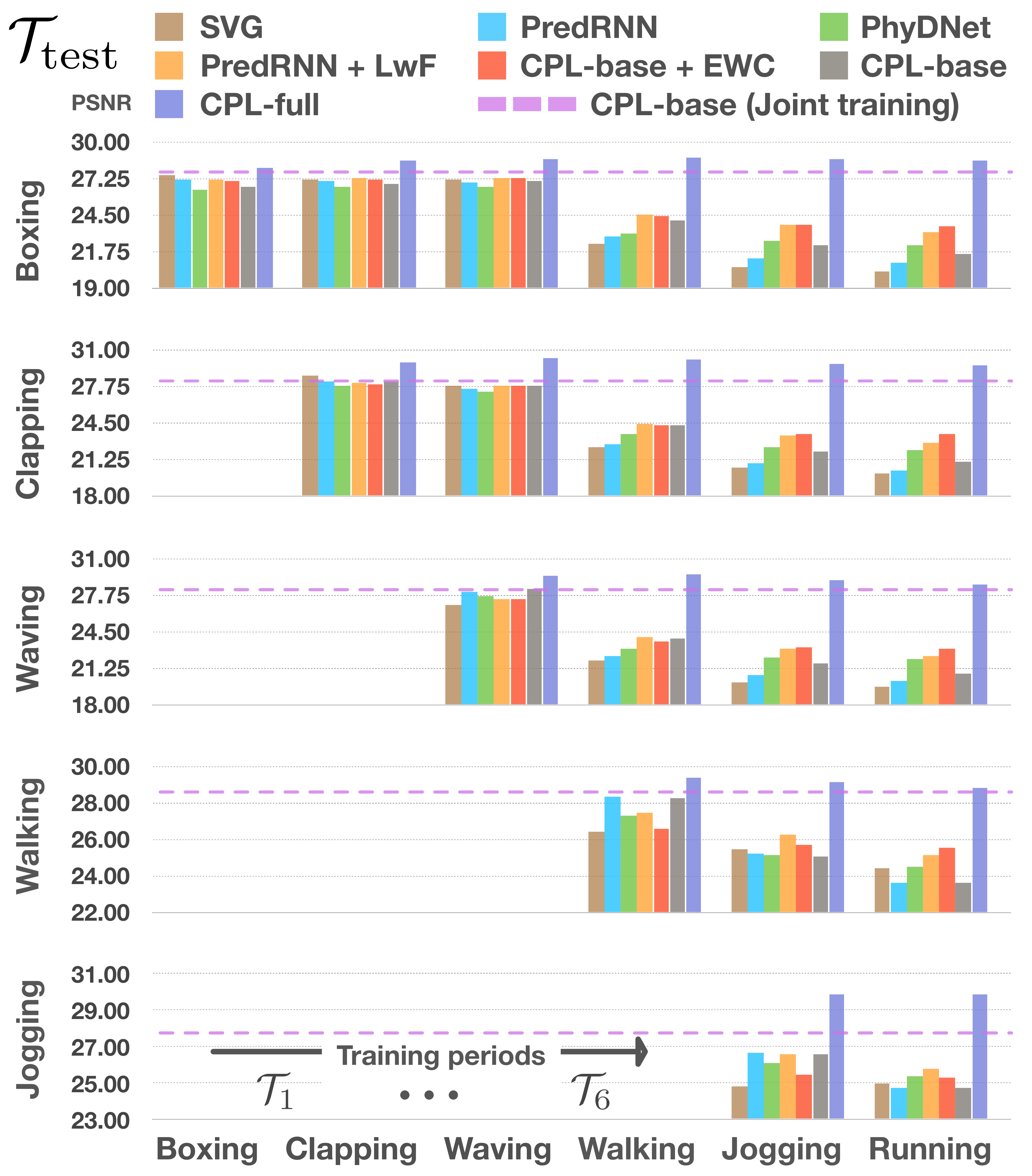}
\vspace{-15pt}
\caption{Results on the KTH benchmark. The horizontal axis represents the sequential training process, and the vertical axes represent test results on particular tasks after each training period.}
\label{fig:kth_psnr_compare}
\end{figure}

\paragraph{Quantitative comparison.} 
\tab{tab:kth} shows the quantitative results on the test sets of all $6$ tasks after the last training period of the models on the last task.
We can observe that CPL significantly outperforms the compared video prediction methods and continual learning methods in both PSNR and SSIM.
Furthermore, an interesting result is that our approach even outperforms the joint training model, as shown in the bottom line in \tab{tab:kth}.
While we do not know the exact reasons, we state two hypotheses that can be investigated in future work.
First, the Gaussian mixture priors enable the world model to better disentangle the representations of visual dynamics learned in different continual learning tasks. 
Second, the predictive experience replay allows the pre-learned knowledge on previous tasks to facilitate the learning process on new tasks.
\fig{fig:kth_psnr_compare} provides the intermediate test results on particular tasks after each training period, which confirm the above conclusions.

\newcommand{\yes}{{\scriptsize\Checkmark}}
\newcommand{\no}{{\scriptsize\XSolidBrush}}

\begin{table}
\small
\centering
\begin{tabular}{|c|c|c|c|cc|}
\hline
Replay & Infer ${k}$ & Random ${k}$ & Adapt & PSNR &  SSIM\\
\hline
\hline
\no & \no & \no & \no & 22.96 & 68.98 \\
\yes & \no & \no & \no & 27.21 & 79.99  \\
\yes & \yes & \no & \no & 27.82 & 81.51 \\
\yes & \no & \yes & \no & 26.56 & 78.64 \\
\yes & \yes & \no & \yes & \textbf{29.12} & \textbf{84.50}    \\
\hline
\end{tabular}
\vspace{-5pt}
\caption{Ablation study for each component of CPL on the KTH benchmark. ``Replay'' denotes the use of predictive experience replay. ``Infer $k$'' indicates the use of non-parametric task inference. ``Random $k$'' means that the world model takes as input a random task label at test time. ``Adapt'' means test-time adaptation.}
\label{tab:Ablations_kth}
\vspace{-5pt}
\end{table}

\begin{figure}[t]
\centering
\includegraphics[width=1.0\linewidth]{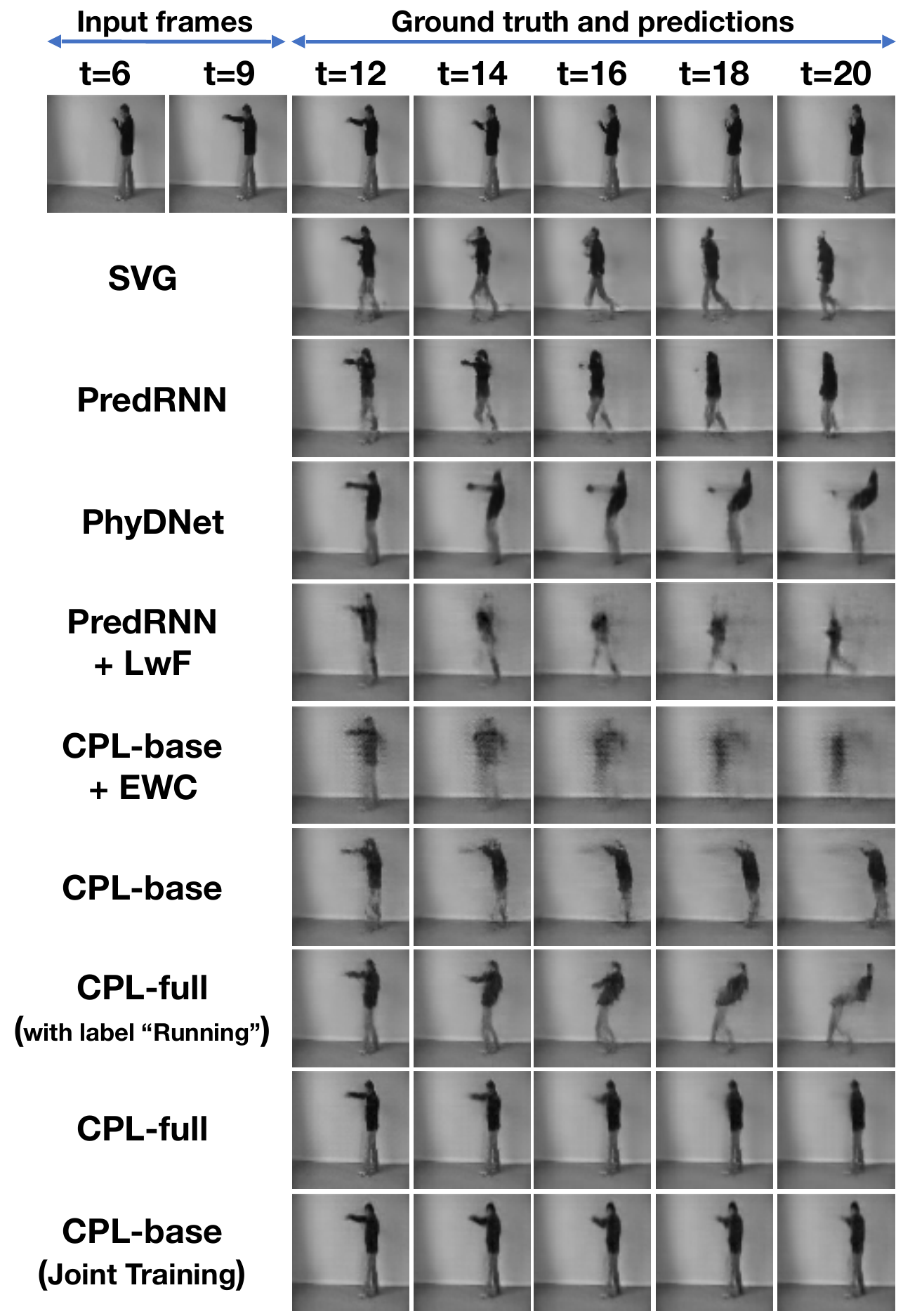}
\vspace{-15pt}
\caption{Showcases of predicted frames of the first task (\ie, \textit{Boxing}) after the training period of the last task (\ie, \textit{Running}).}
\vspace{-10pt}
\label{fig:kth_qualitative}
\end{figure}

\myparagraph{Qualitative comparison.}
We visualize a sequence of predicted frames on the first task of KTH in \fig{fig:kth_qualitative}. 
As shown, all existing video prediction models and even the one with LwF generate future frames with the dynamics learned in the last task (\ie, \textit{Running}), which clearly demonstrates the influence of the dynamics shift. 
Images generated by CPL-base+EWC suffer from a severe blur effect, indicating that the model cannot learn disentangled representations for different dynamics in the non-stationary training environments.
In comparison, CPL produces more reasonable results.
%
To testify the necessity of task inference, we also provide incorrect task labels for CPL.
As shown in the third line from the bottom, the model takes as input the \textit{Boxing} frames along with an erroneous task label of \textit{Running}.
Interestingly, CPL combines the inherent dynamics of input frames (reflected in motion of arms) with the dynamics priors from the input task label (reflected in motion of legs).

\subsection{Ablation Study}
\label{sec:ablation}
\paragraph{Effectiveness of each component in CPL.}
We conduct ablation studies on the KTH benchmark step by step. 
In \tab{tab:Ablations_kth}, the first line shows the results of the CPL-base model and the bottom line corresponds to our final approach. 
In the second line, we train CPL-base with predictive experience replay and observe a significant improvement from $22.96$ to $27.21$ in PSNR.
In the third line, we improve the world model with mixture-of-Gaussian priors and accordingly perform non-parametric task inference at test time. We observe consistent improvements in both PSNR and SSIM upon the previous version of the model.
In the fourth line, we skip the non-parametric task inference during testing and use a random task label instead. We observe that the performance drops from $27.82$ to $26.56$ in PSNR, indicating the importance of task inference to predictive experience replay.
Finally, in the bottom line, we introduce the self-supervised test-time adaptation. It shows a remarkable performance boost compared with all the above variants.

\myparagraph{Is CPL robust to the task order?}
As shown in \tab{tab:kth_false_label}, we further conduct experiments to analyze that whether CPL can effectively alleviate catastrophic forgetting regardless of the task order. We additionally train the CPL model in $3$-$4$ random task orders.
From the results, we find that the proposed techniques including mixture world model, predictive experience replay, and non-parametric task inference are still effective despite the change of training order.

\begin{table}[t]
\small
    \centering
    \begin{tabular}{|l|cc|}
    \hline
    Dataset       & PSNR & SSIM ($\times 10^{-2}$) \\
    \hline
    \hline
    
    RoboNet     & 23.58 $\pm$ 0.28    &  79.67 $\pm$ 3.75                         \\
    KTH         &   28.93 $\pm$ 0.14 &  83.99 $\pm$ 0.40           \\
    \hline
\end{tabular}
\vspace{-5pt}
\caption{Robustness of CPL on random task orders.}
	\label{tab:kth_false_label}
	\vspace{-10pt}
\end{table}

\section{Related Work}
\paragraph{Continual learning of supervised tasks.} 
Continual learning is designed to cope with the continuous information flow, retaining or even optimizing old knowledge while absorbing new knowledge.
Mainstream paradigms include regularization, replay, and parameter isolation~\cite{delange2021continual}. 
The regularization approaches typically tackle catastrophic forgetting~\cite{goodfellow2013empirical} by constraining the learned parameters on new tasks with additional loss terms, \eg, EWC~\cite{kirkpatrick2017overcoming}, or distilling knowledge from old tasks, \eg, LwF~\cite{li2017learning}.
For replay-based approaches, a typical solution is to retain a buffer on earlier tasks of representative data or feature exemplars~\cite{rebuffi2017icarl,riemer2019learning,ayub2021eec}. Some approaches also use generative networks to encode the previous data distribution and synthesize fictitious data for experience replay, \eg, DGR~\cite{shin2017continual} and CURL~\cite{rao2019continual}.
The parameter isolation approaches allow the neural networks to dynamically expand when new tasks arrive~\cite{rusu2016progressive} or encourage the new tasks to use previously ``unused'' parameter subspaces~\cite{he2018overcoming}.

\myparagraph{Continual learning of unsupervised tasks.} 
Most existing approaches are mainly focused on supervised tasks of image data.
Despite the previous literature that discussed unsupervised CL \cite{rao2019continual,cha2021co2l,ke2021classic}, our approach is significantly different from these methods as it explores the specific challenges of continual predictive learning for video data, especially the covariate-dynamics-target shift.
The most related method to CPL is CURL \cite{rao2019continual}, which introduces a mixture-of-Gaussian latent space for class-incremental CL and combat forgetting via generative replay. 
There are three major differences between CPL and CURL. First, CURL cannot be directly used in our setup as it does not handle the dynamics shift in non-stationary spacetime, while CPL tackles it through a new world model. 
Second, CPL greatly benefits from the carefully-designed predictive replay algorithm, while it is extremely difficult for CURL to replay valid video frames using a fully generative model alone.
Third, CPL provides a non-parametric task inference method as opposed to the model-based inference method in CURL.

\myparagraph{Video prediction.}
RNN-based models have been widely used for deterministic video prediction \cite{ranzato2014video, srivastava2015unsupervised,shi2015convolutional,de2016dynamic,wang2017predrnn,wang2019eidetic,su2020convolutional, yao2020unsupervised}.
Shi \etal \cite{shi2015convolutional} proposed ConvLSTM to improve the learning ability of spatial information by combining convolutions with LSTM transitions. 
Following this line, Wang \etal \cite{wang2017predrnn} proposed PredRNN, modeling memory cells in a unified spatial and temporal representation.
Stochastic video prediction models assume that different plausible outcomes would be equally probable for the same input, and thus incorporate uncertainty in the models using GANs \cite{vondrick2016generating,tulyakov2018mocogan} or  VAEs \cite{denton2018stochastic,babaeizadeh2018stochastic, lee2018stochastic, castrejon2019improved, franceschi2020stochastic}.
Particularly, Yao \etal proposed to adapt video prediction models from multiple source domains to a target domain via distillation~\cite{yao2020unsupervised}. 
However, it cannot be easily used as a solution to continual predictive learning, as the number of retained model parameters increases linearly with the number of tasks.
\section{Discussion}

In this paper, we explored a new research problem of continual predictive learning, which is challenging due to the co-existence of the covariate, dynamics, and target shift.
We proposed an approach named CPL, whose major contributions of CPL can be viewed in three aspects.
First, it presents a new world model to capture task-specific visual dynamics in a Gaussian mixture latent space.
Second, it introduces the predictive experience replay method to overcome the forgetting issue in the world model. 
Third. it leverages a non-parametric task inference strategy to avoid coupling the forgetting issues caused by the introduction of a task inference model. 
Our approach has shown competitive results on RoboNet and KTH benchmarks, achieving remarkable improvements over the na\"ive combinations of existing world models and CL algorithms.

Although CPL can be easily extended to more complex tasks, the potential limitation is that it has not been evaluated in the entire pipeline of vision-based robot control, which includes the processes of predictive learning and decision making.
In future work, we plan to integrate CPL in a model-based reinforcement learning framework to further validate its effectiveness for downstream tasks. 

\section{Acknowledgement}
This work was supported by NSFC grants (U19B2035, 62106144, 62021002, 62022050), Shanghai Municipal Science and Technology Major Project (2021SHZDZX0102), and Shanghai Sailing Program (21Z510202133).

{\small
\bibliographystyle{latex/ieee_fullname}
\bibliography{latex/egbib}
}

\end{document}



\title{Supplementary Material\\ \textit{Continual Predictive Learning from Videos}}

\author{First Author\\
Institution1\\
Institution1 address\\
{\tt\small firstauthor@i1.org}
\and
Second Author\\
Institution2\\
First line of institution2 address\\
{\tt\small secondauthor@i2.org}
}
\maketitle

\noindent\begin{large}\textbf{0. Summary of The Supplementary Material}\end{large}

\begin{enumerate}
    \item Model details of the proposed CPL approach.
    \myitem More experimental configurations and training details.
    \myitem Full quantitative comparisons on RoboNet.
    \myitem Robustness analyses to the training order on RoboNet.
    \myitem Further qualitative results on RobotNet in both action-free and action-conditioned setups.
    \myitem Qualitative results for all previous tasks on KTH, which show significant improvements over the prior art.
\end{enumerate}

\section{Model Architecture Details}
\begin{figure}[h]
\centering
\includegraphics[width=0.95\linewidth]{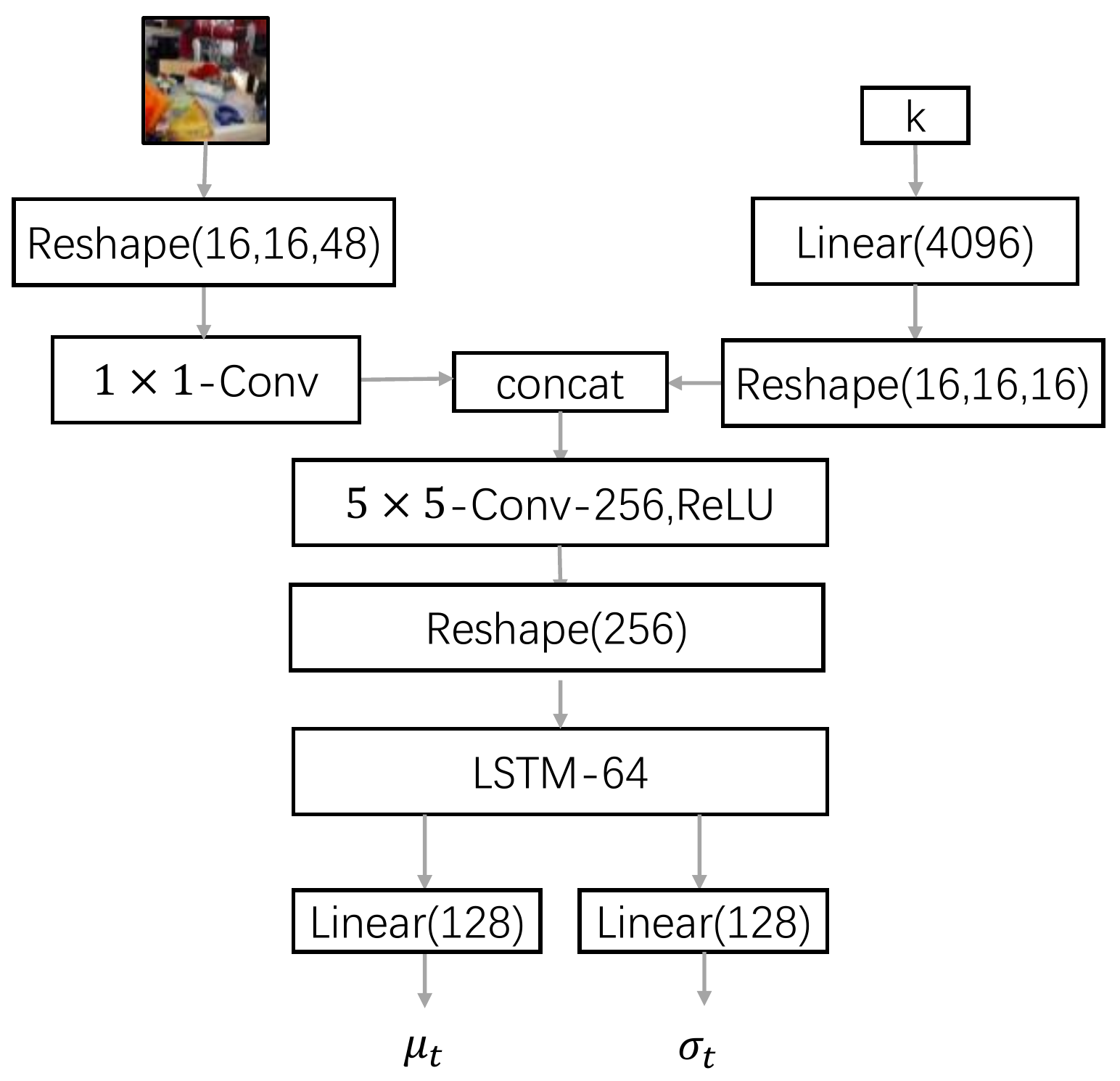}
\caption{Architecture details of the encoding module and the representation module in our Mixture World Model.
}
\label{fig:final_E}
\end{figure}

\begin{figure}[h]
\centering
\includegraphics[width=0.95\linewidth]{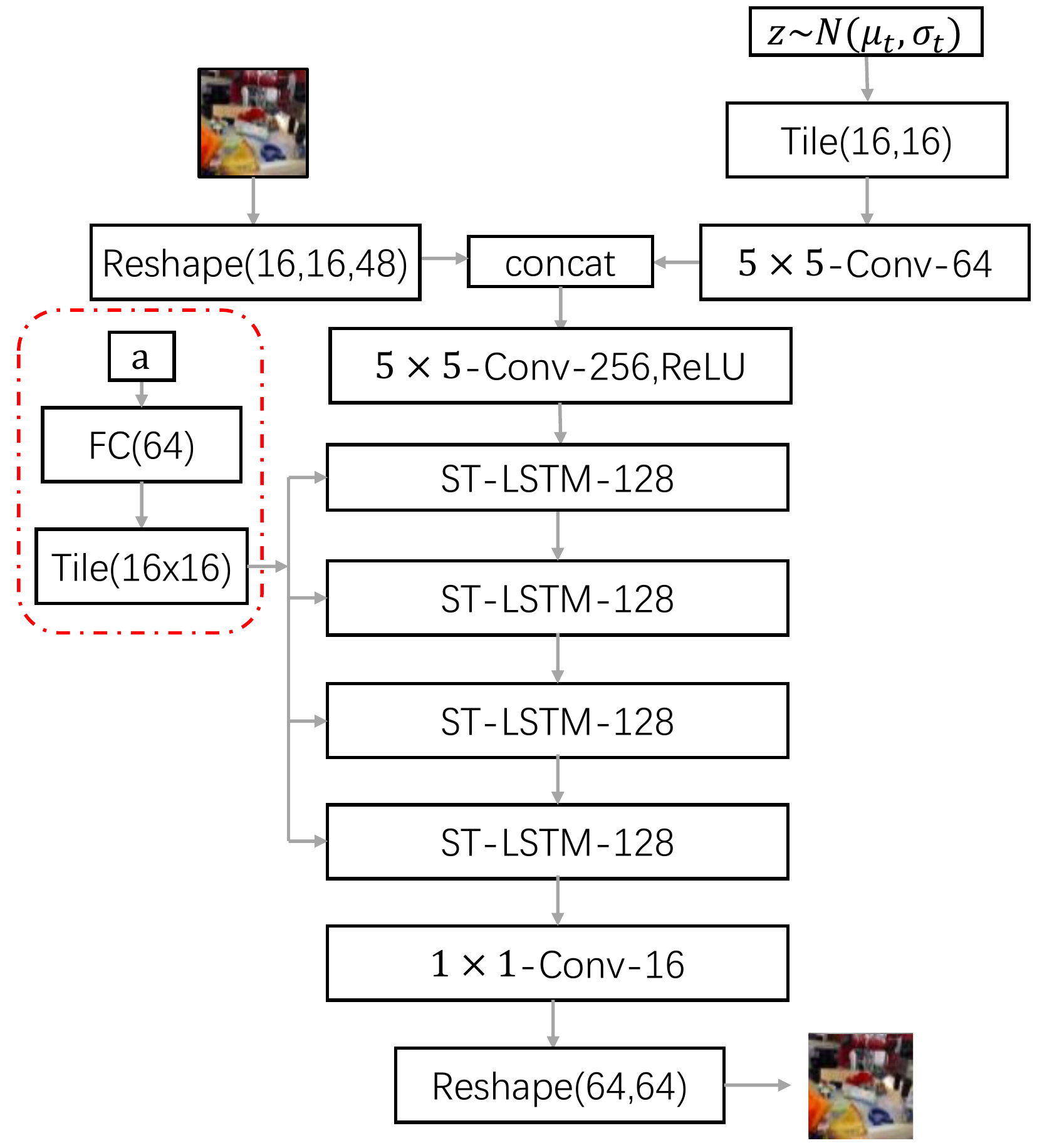}
\caption{Architecture details of the dynamic module in the proposed Mixture World Model. Modules in the red dashed box are only used when actions are provided.
}
\label{fig:final_D}
\end{figure}

\begin{figure}
\centering
\includegraphics[width=1.0\linewidth]{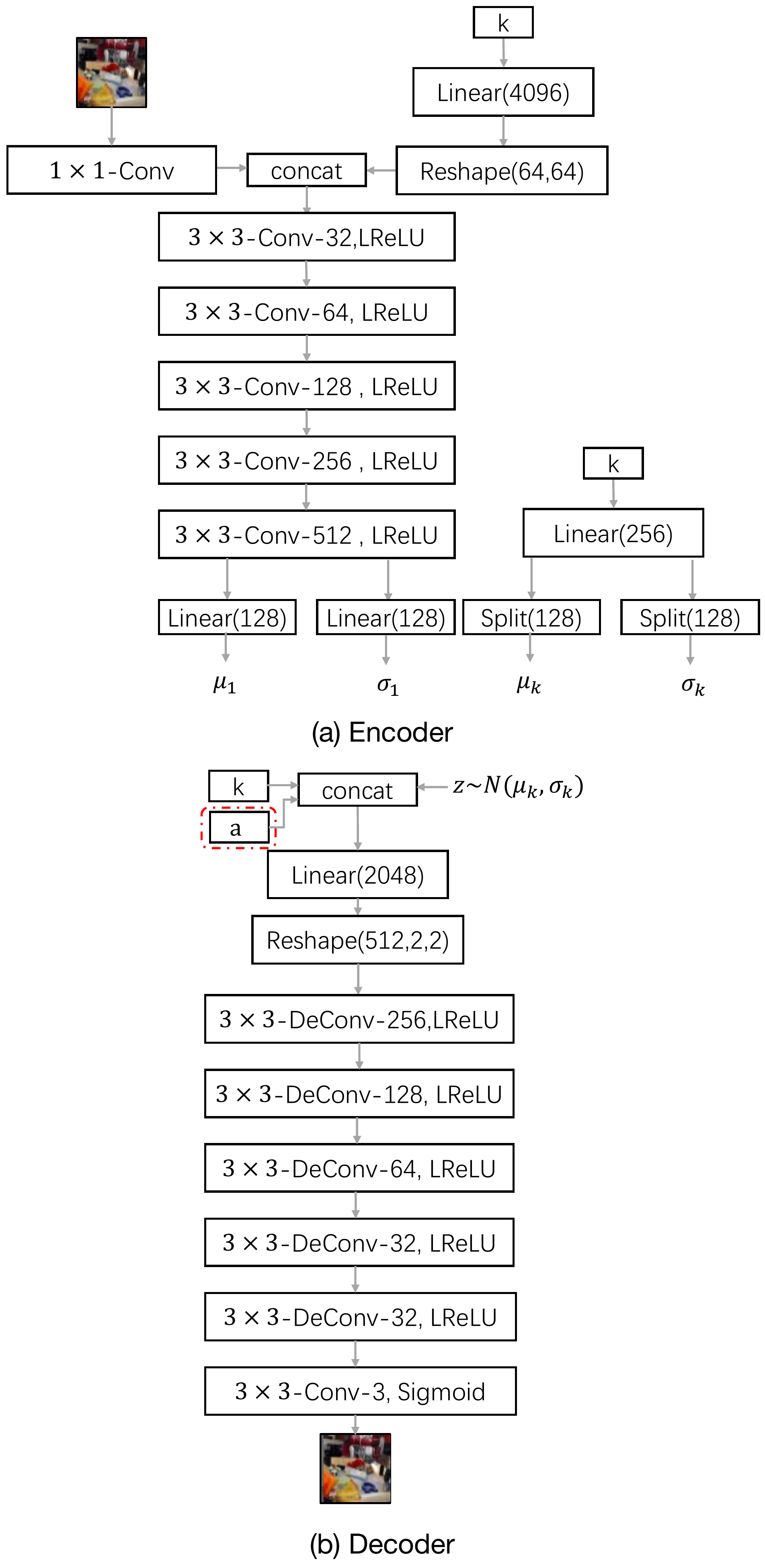}
\caption{Architecture of the generative model in the proposed Predictive Experience Replay scheme, which learns to generate the initial frames of previous tasks. Modules in the red dashed box are only used when actions are provided.
}
\label{fig:final_replay}
\end{figure}

\fig{fig:final_E} and \fig{fig:final_D} provide the detailed model architectures of our Mixture World Model.
\fig{fig:final_replay} shows the details of the generative model that generates the initial frames for Predictive Experience Replay.

\section{Experimental Configurations}

We here provide the training details of CPL. 
In the predictive experience replay scheme, the number of rehearsal video sequences from all previous tasks is about one-third of that used in the current task. All models are trained using the Adam optimizer with $\beta_{1}=0.9$ and $\beta_{2}=0.999$, and the learning rate is set to $0.0005$ for the KTH benchmark and $0.0001$ for RoboNet. Besides, the mini-batch size is set to $32$ for KTH and $16$ for RoboNet. The input frames are pre-resized to $64\times 64$ for both benchmarks. We optimize the entire model by $30{,}000$ iterations for each task in the continual learning process. We train all compared models on a GTX 3090 GPU.

\section{Further Quantitative Results on RoboNet}

\begin{figure}[t]
\centering
\includegraphics[width=1.0\linewidth]{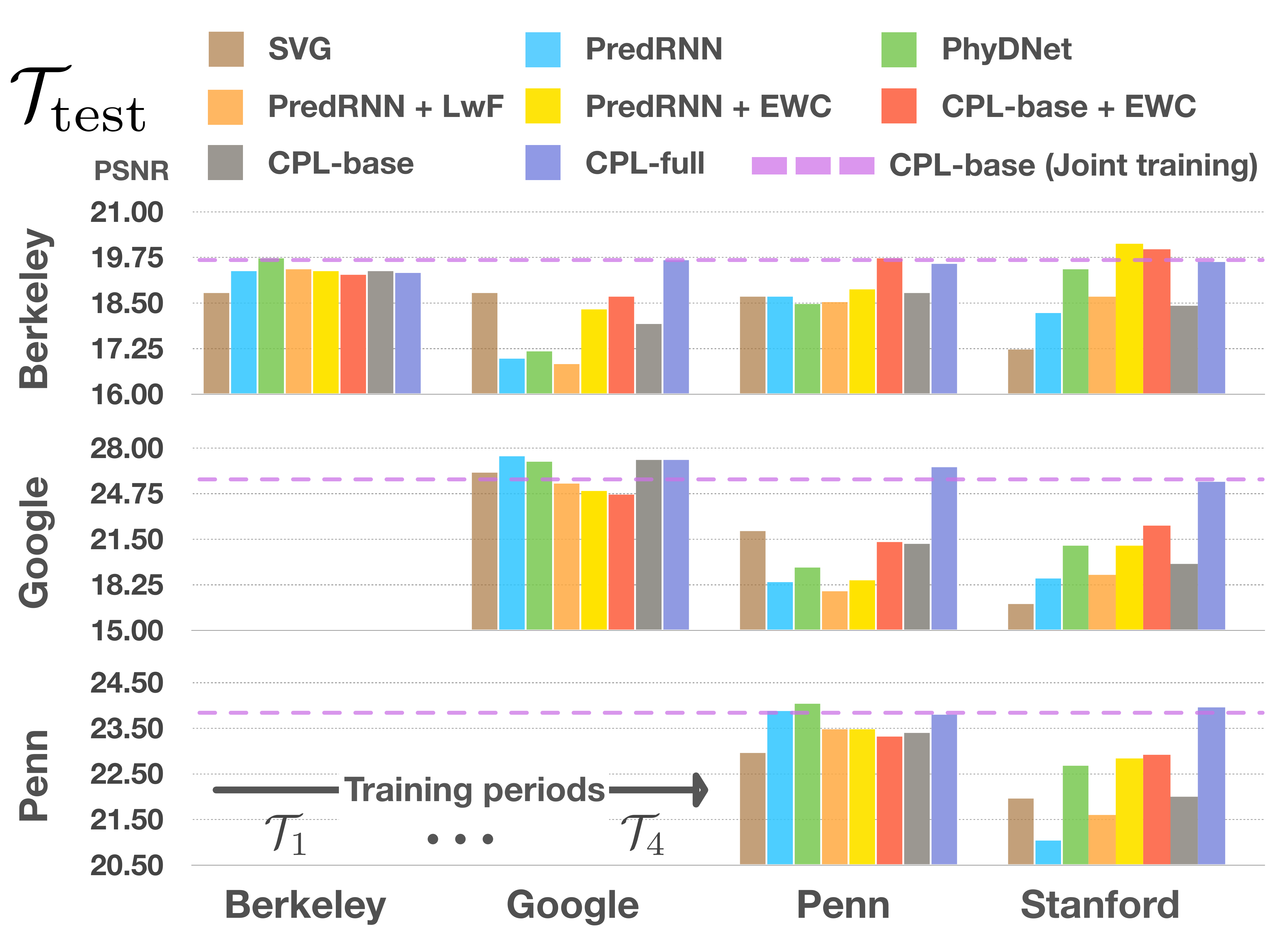}
\caption{Results on the action-free RoboNet benchmark.
}
\label{fig:robonet_free_full_cond}
\end{figure}

\begin{figure}[t]
\centering
\includegraphics[width=1.0\linewidth]{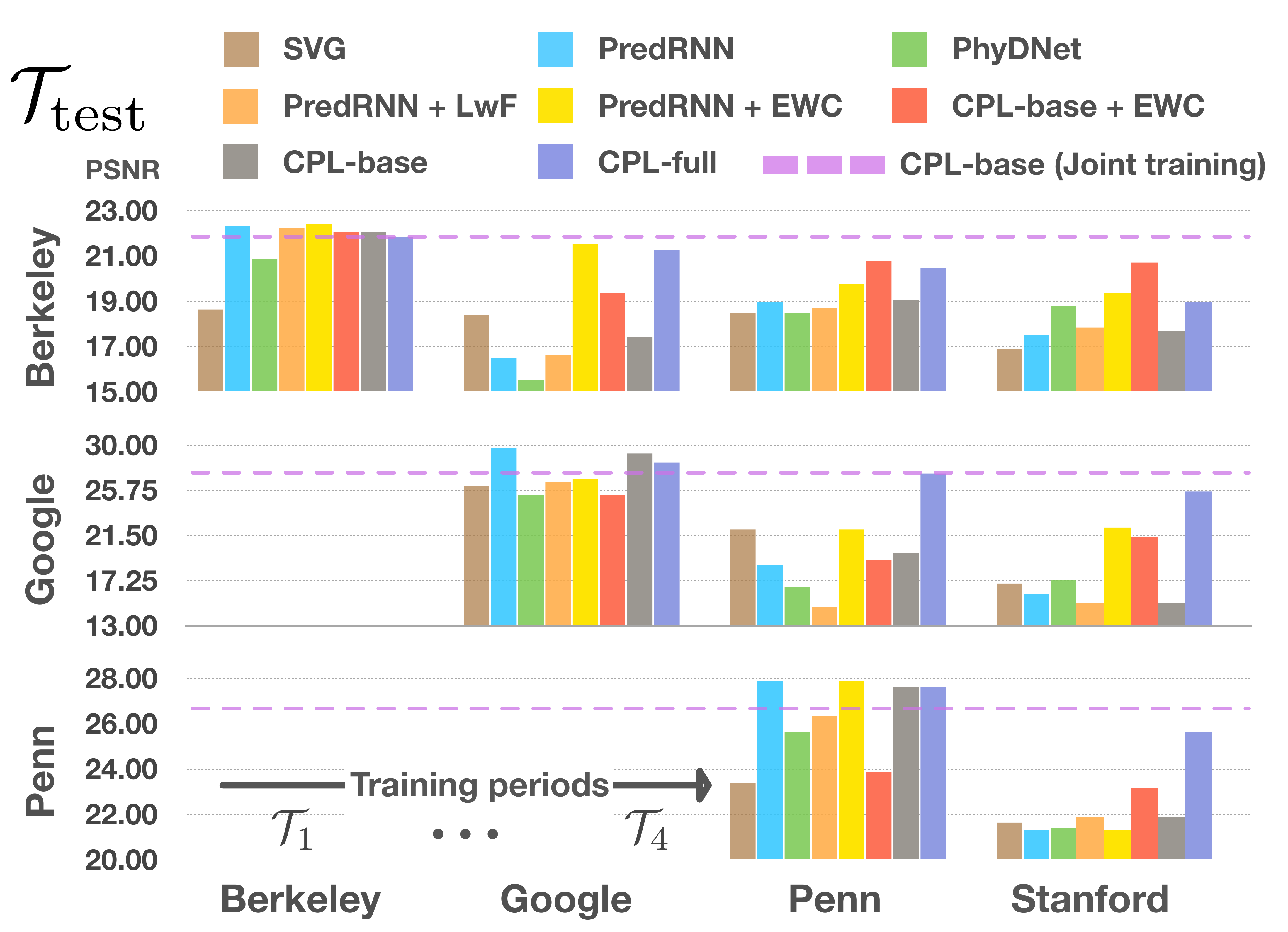}
\caption{Results on the action-conditioned RoboNet benchmark.
}
\label{fig:robonet_conditioned_full_cond}
\end{figure}

\fig{fig:robonet_free_full_cond} shows the full quantitative comparisons on particular tasks after individual training periods on the action-free RoboNet benchmark. 
\fig{fig:robonet_conditioned_full_cond} shows the corresponding results under the action-conditioned setup.

\section{Robustness of CPL to RoboNet Task Order}
\label{sec:order}
As shown in \tab{tab:robonet_actionfree} and \tab{tab:robonet_action}, we further conduct experiments on RoboNet to analyze that whether CPL can effectively alleviate catastrophic forgetting regardless of the task order.
Specifically, we train CPL with a task order of \textit{Penn} $\rightarrow$ \textit{Google} $\rightarrow$ \textit{Berkeley} $\rightarrow$ \textit{Stanford}, which is different from what has been used in \underline{Table $1$} in the manuscript. 
From the results, we find that the proposed techniques, \ie, (i) the mixture world model, (ii) the predictive experience replay, and (iii) the non-parametric task inference, are still effective despite the change of training order under both action-conditioned and action-free setups.

\section{Further Qualitative Results on RoboNet}

\subsection{Action-Free Video Prediction}

\begin{figure}[h]
\centering
\includegraphics[width=1.0\linewidth]{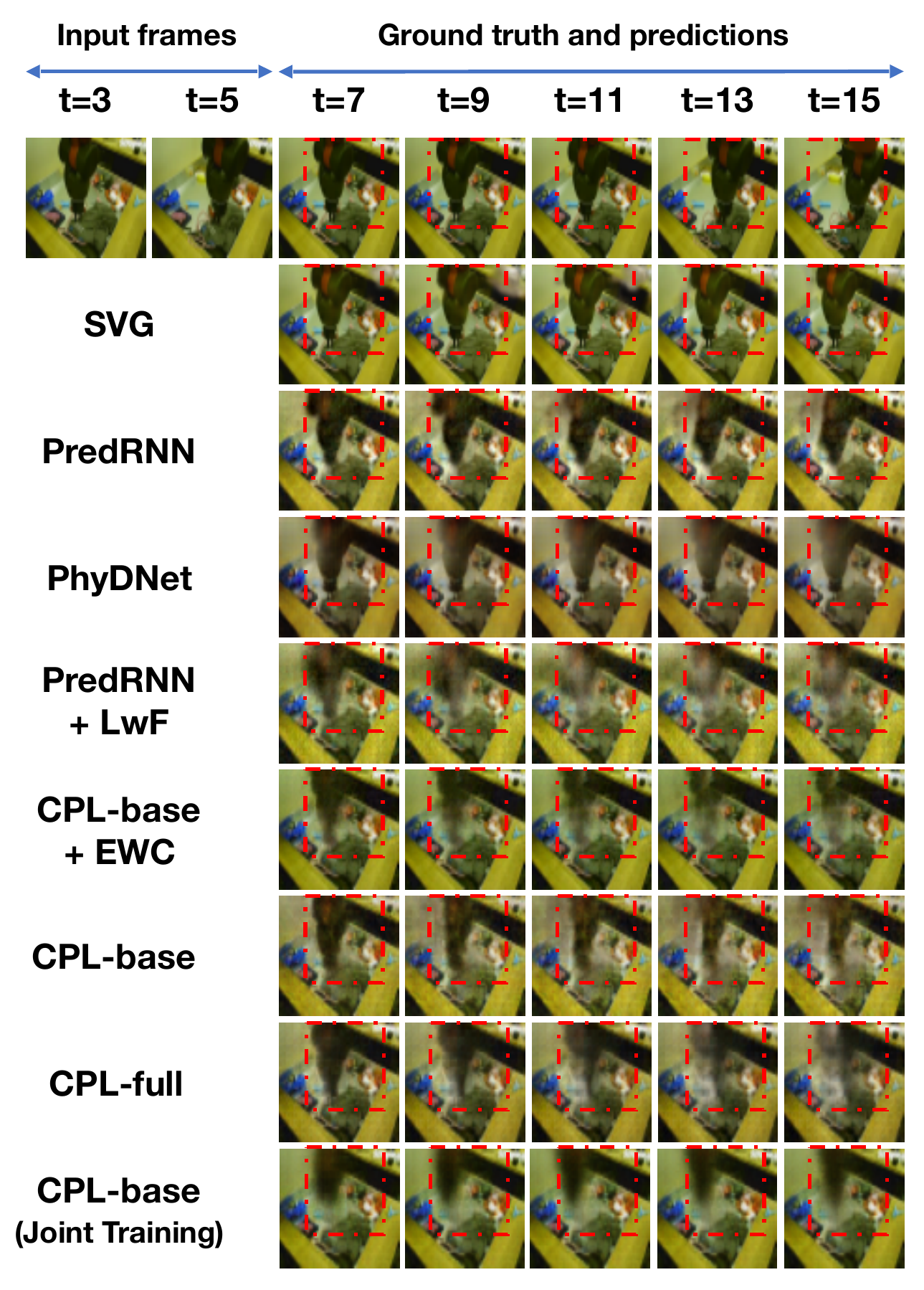}
\caption{Showcases of predicted frames on the action-free RoboNet benchmark.
The example is from the \textbf{\textit{Penn}} environment, which is the \textbf{first} task in the process of continual learning. 
We use models that have finished the last training period on \textit{Stanford}.
}
\label{fig:robonet_free_full}
\end{figure}

\fig{fig:robonet_free_full} gives examples for predicted frames on RoboNet under the action-free setup.  
We here follow the task order described in the above section, \ie, \textit{Penn} $\rightarrow$ \textit{Google} $\rightarrow$ \textit{Berkeley} $\rightarrow$ \textit{Stanford}.
The input sequence is randomly sampled from the test set of the first environment (\textit{Penn}).
The prediction results are made by models that have finished the entire training procedure (\ie, after the last training period on \textit{Stanford}).

\begin{table}[t]
    \centering
    \begin{tabular}{|l|cc|}
    \hline
    Method       & PSNR & SSIM ($\times 10^{-2}$) \\
    \hline
    \hline
    
    CPL-base     & 19.71 $\pm$0.01    &  68.37 $\pm$0.04                         \\
    CPL-full         &   \textbf{22.07$\pm$0.04}  &  \textbf{77.08 $\pm$0.14}           \\
    \hline
\end{tabular}
\caption{Results on action-free RoboNet by models trained with a different task order.}
	\label{tab:robonet_actionfree}
\end{table}

\begin{table}[t]
    \centering
    \begin{tabular}{|l|cc|}
    \hline
    Method       & PSNR & SSIM ($\times 10^{-2}$) \\
    \hline
    \hline
    
    CPL-base     & 19.07 $\pm$0.00    &  62.56 $\pm$0.02                         \\
    CPL-full         &   \textbf{23.22$\pm$0.02}  &  \textbf{71.32 $\pm$0.15}           \\
    \hline
\end{tabular}
\caption{Results on action-conditioned RoboNet by models trained with a different task order.}
	\label{tab:robonet_action}
\end{table}

As we can see from this figure, all existing video prediction models, including SVG, PredRNN, and PhyDNet, do not have an accurate prediction of the motion of the robot arm.
Even the Joint-Training baseline (see the bottom line) tends to produce rather static images across multiple time steps.
Compared with the models based on LwF and EWC, our approach (CPL-full) makes less blurry predictions around the robot arm in future frames.

\begin{figure*}[t]
\centering
\includegraphics[width=1.0\linewidth]{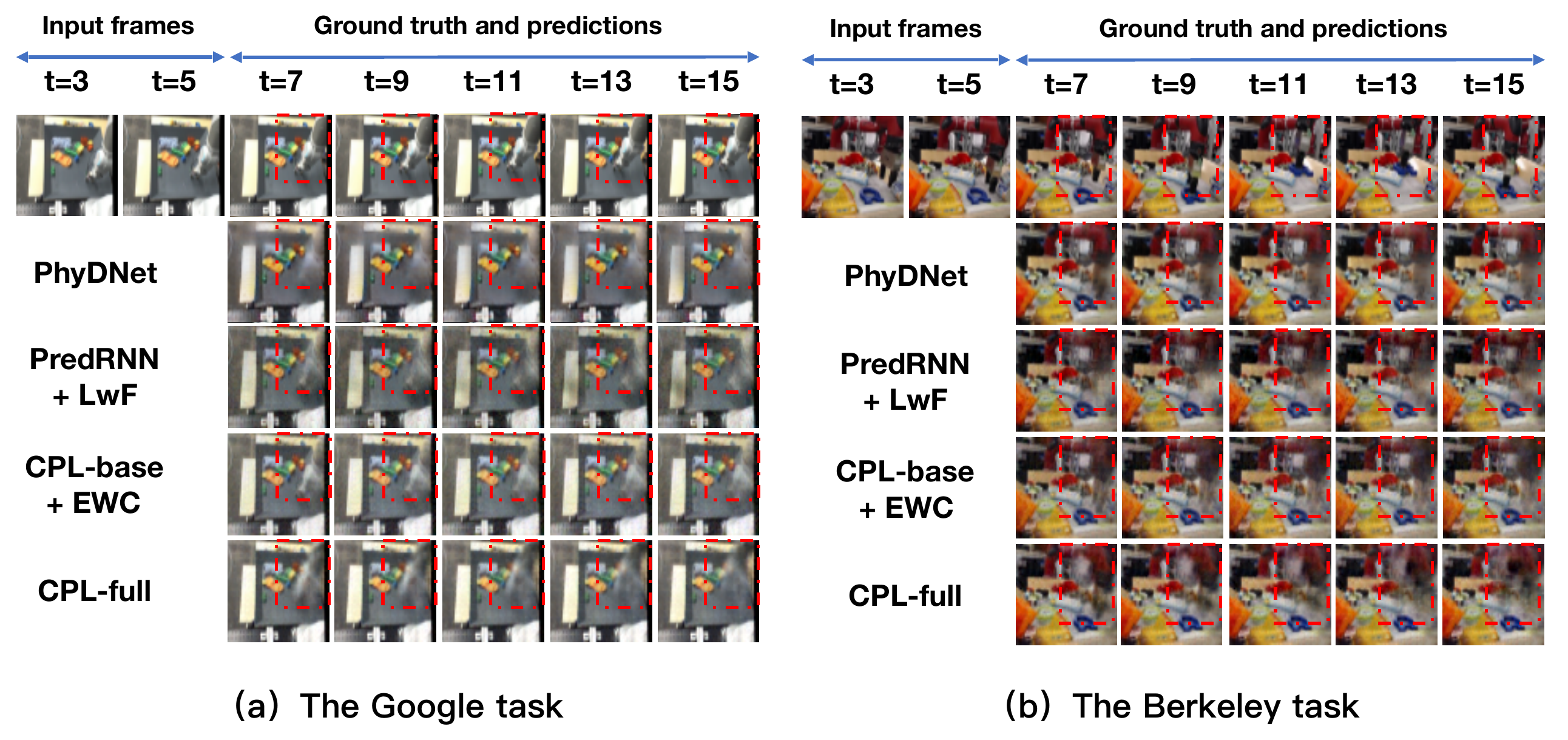}
\caption{Action-free prediction results from (a) the \textbf{\textit{Google}} environment and (b) the \textbf{\textit{Berkeley}} environment, which are respectively the \textbf{second} and the \textbf{third} task in continual learning setup on RoboNet. 
The task order at training time is \textit{Penn} $\rightarrow$ \textit{Google} $\rightarrow$ \textit{Berkeley} $\rightarrow$ \textit{Stanford}.
We use models that have finished the last training period on \textit{Stanford}.
For the first task of \textit{Penn}, please refer to \fig{fig:robonet_free_full}. 
}
\label{fig:robonet_free_G_B}
\end{figure*}

In \fig{fig:robonet_free_G_B}, we also provide results on the other two previous tasks, \ie, \textit{Google} and \textit{Berkeley}.
As above, the predicted frames are generated by models that have finished the last training period on \textit{Stanford}.

\subsection{Action-Conditioned Video Prediction}
\begin{figure*}[t]
\centering
\includegraphics[width=1.0\linewidth]{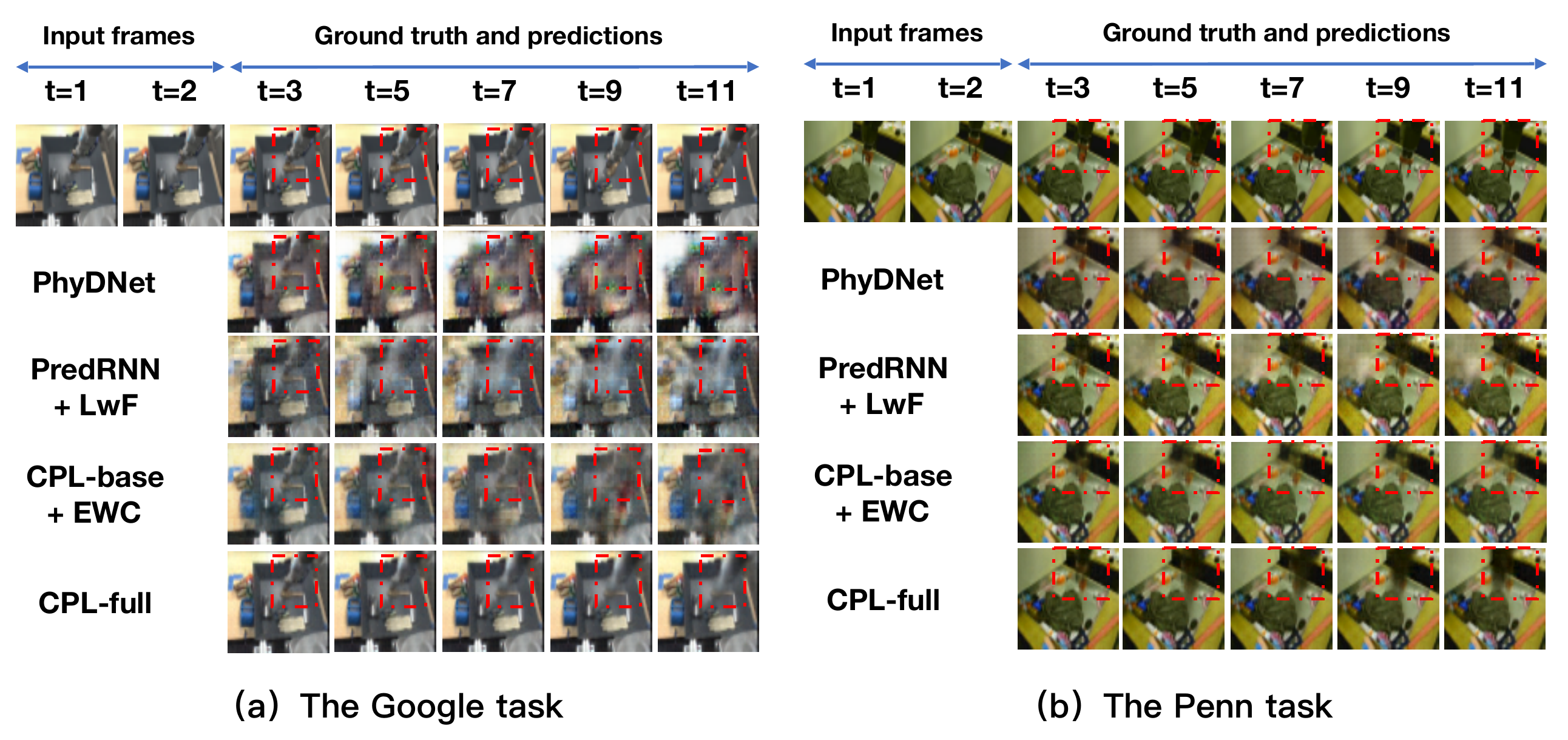}
\caption{
Action-conditioned prediction results from (a) the \textbf{\textit{Google}} environment and (b) the \textbf{\textit{Penn}} environment, which are respectively the \textbf{second} and the \textbf{third} task in continual learning setup on RoboNet. 
The task order at training time is \textit{Berkeley} $\rightarrow$ \textit{Google} $\rightarrow$ \textit{Penn} $\rightarrow$ \textit{Stanford}.
We use models that have finished the last training period on \textit{Stanford}. For the first task of \textit{Berkeley}, please refer to \underline{Fig. $4$} in the manuscript.
}
\label{fig:robonet_conditioned_G_P}
\end{figure*}

For the action-conditioned setup, besides the results shown \underline{Fig. $4$} in the manuscript, we here provide results on other two tasks (\ie, \textit{Google} and \textit{Penn}) in \fig{fig:robonet_conditioned_G_P}.
To be consistent with the results in the manuscript, we still follow the training order of \textit{Berkeley} $\rightarrow$ \textit{Google} $\rightarrow$ \textit{Penn} $\rightarrow$ \textit{Stanford}, and use the final models after the last training period on \textit{Stanford}.

As we can see, our approach (CPL-full) shows sharper and more accurate prediction results than the state-of-the-art video prediction model (\ie, PhyDNet), as well as the continual learning methods (\ie, LwF and EWC).

\section{Further Qualitative Results on KTH}

\begin{figure*}[t]
\centering
\includegraphics[width=1.0\linewidth]{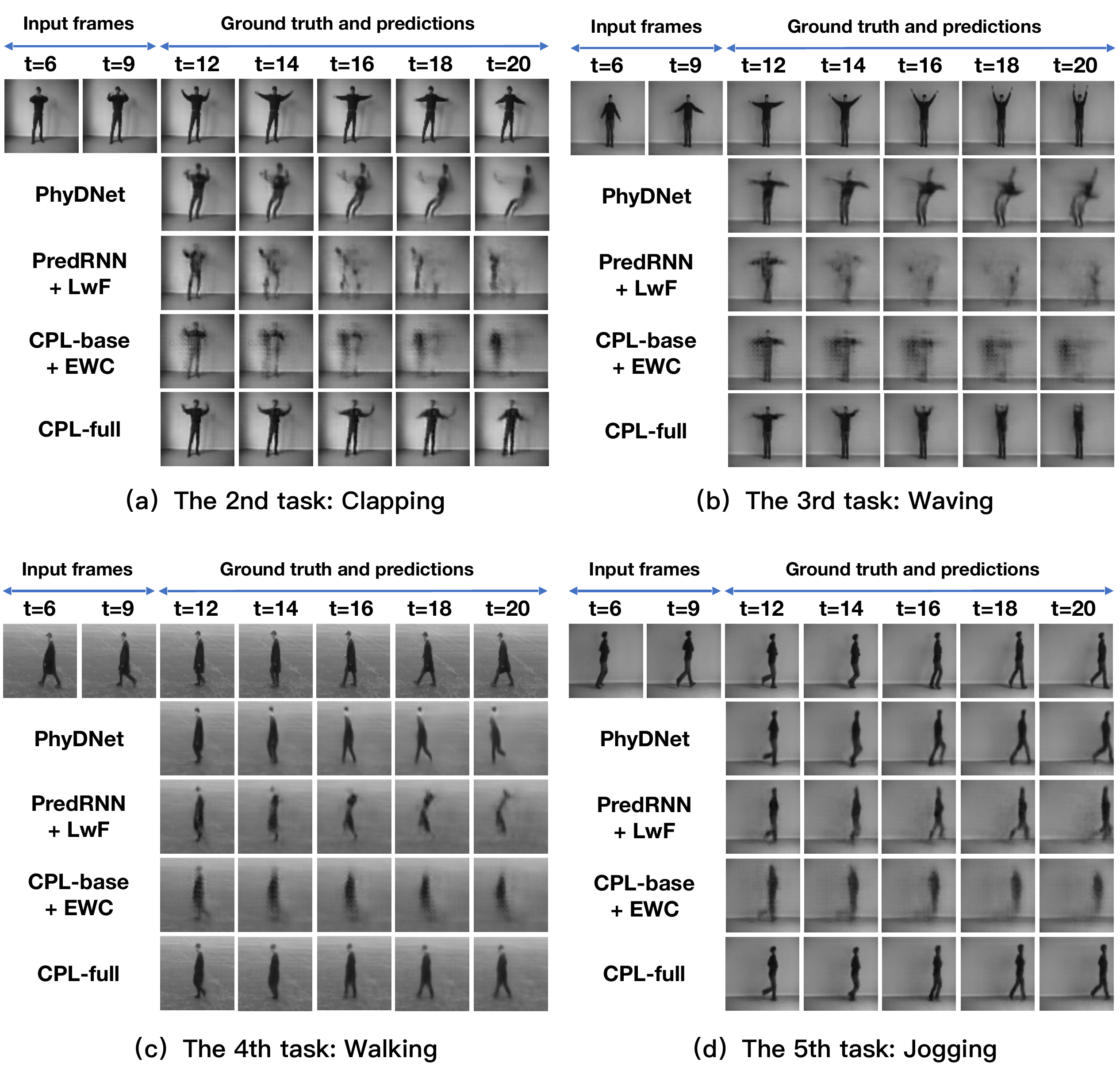}
\caption{
Video prediction results on the previous tasks before \textit{Running}. We use models that have finished the last training period on \textit{Running}. For the first task of \textit{Boxing}, please refer to \underline{Fig. $6$} in the manuscript. Note that our CPL approach is the only one that shows the ability to effectively mitigate catastrophic forgetting and generate \textbf{CORRECT} motions from corresponding observation frames. 
}
\label{fig:kth_all}
\end{figure*}

On the KTH benchmark, we set the the training order as \textit{Boxing} $\rightarrow$ \textit{Clapping} $\rightarrow$ \textit{Waving} $\rightarrow$ \textit{Walking} $\rightarrow$ \textit{Jogging} $\rightarrow$ \textit{Running}.
In \underline{Fig. $6$} in the manuscript, we have provided the prediction showcases from the test set of the first task (\ie, \textit{Boxing}) by models trained on the last task (\ie, \textit{Running}).
Here, we show corresponding results on the other four previous tasks in \fig{fig:kth_all}.


Our approach shows remarkable improvements over the state-of-the-art video prediction model (\ie, PhyDNet), as well as the continual learning methods (\ie, LwF and EWC).
Notably, none of the compared models can  ``remember'' the motion on previous tasks, especially for the \textit{Clapping} and \textit{Waving} tasks that have long gone.
While our CPL approach is the only one that shows the ability to effectively mitigate catastrophic forgetting and generate \textbf{CORRECT} motions from corresponding observation frames.

\clearpage


\clearpage
